\pdfoutput=1
\documentclass{article}

\usepackage[utf8]{inputenc} 
\usepackage[T1]{fontenc}    
\usepackage{hyperref}       
\usepackage{url}            
\usepackage{booktabs}       
\usepackage{amsfonts}       
\usepackage{nicefrac}       
\usepackage{microtype}      
\usepackage{xcolor}         
\usepackage{xspace}
\usepackage{adjustbox}
\usepackage{amsthm}
\usepackage{textcomp}
\usepackage{siunitx}
\usepackage{color}
\usepackage{hyperref}
\usepackage{xcolor}
\usepackage[font={small}]{caption}
\usepackage{epstopdf}
\usepackage{graphicx}
\usepackage{caption}
\usepackage{verbatim} 
\usepackage{bbm}
\usepackage{mathtools}
\usepackage{enumitem}
\usepackage{subfigure}
\usepackage{multirow}
\usepackage{cases}
\usepackage{kantlipsum}
\allowdisplaybreaks

\DeclarePairedDelimiter\floor{\lfloor}{\rfloor}

\usepackage{listings}
\definecolor{codegreen}{rgb}{0,0.6,0}
\definecolor{codegray}{rgb}{0.5,0.5,0.5}
\definecolor{codepurple}{rgb}{0.58,0,0.82}
\definecolor{backcolour}{rgb}{0.95,0.95,0.92}
\lstdefinestyle{mystyle}{
    inputencoding=utf8,
    extendedchars=true,
    commentstyle=\color{codegreen},
    keywordstyle=\color{magenta},
    numberstyle=\tiny\color{codegray},
    stringstyle=\color{codepurple},
    basicstyle=\ttfamily\small,
    breakatwhitespace=false,         
    breaklines=true,                 
    captionpos=b,                    
    keepspaces=true,                 
    numbers=left,                    
    numbersep=5pt,                  
    showspaces=false,                
    showstringspaces=false,
    showtabs=false,                  
    tabsize=4,
    literate={λ}{{$\lambda$}}1
}
\newtheorem{theorem}{Theorem}
\newtheorem{lemma}{Lemma}

\newtheorem{remark}{Remark}
\newtheorem*{theorem*}{Theorem}

\newtheorem{assumption}{Assumption}

\theoremstyle{definition}

\theoremstyle{remark}

\usepackage{algorithm}
\usepackage{algorithmic}
\newcommand{\scheme}{\texttt{LightSecAgg}\xspace}
\newcommand{\google}{\texttt{SecAgg}\xspace}
\newcommand{\googlep}{\texttt{SecAgg+}\xspace}
\newcommand{\turbo}{\texttt{TurboAgg}\xspace}
\newcommand{\fast}{\texttt{FastSecAgg}\xspace}
\newcommand{\FedBuff}{\texttt{FedBuff}\xspace}

\usepackage[accepted]{mlsys2022}

\mlsystitlerunning{\scheme: a Lightweight and Versatile Design for \\ Secure Aggregation in Federated Learning}

\begin{document}

\twocolumn[
\mlsystitle{\scheme: a Lightweight and Versatile Design for \\ Secure Aggregation in Federated Learning}




\mlsyssetsymbol{equal}{*}

\begin{mlsysauthorlist}
\mlsysauthor{Jinhyun So}{equal,usc}
\mlsysauthor{Chaoyang He}{equal,usc}
\mlsysauthor{Chien-Sheng Yang}{equal,media}
\mlsysauthor{Songze Li}{hkust,hkust2}
\mlsysauthor{Qian Yu}{princeton}
\mlsysauthor{Ramy E. Ali}{usc}
\mlsysauthor{Basak Guler}{ucr}
\mlsysauthor{Salman Avestimehr}{usc}
\end{mlsysauthorlist}

\mlsysaffiliation{usc}{University of Southern California, Los Angeles, California, USA}
\mlsysaffiliation{media}{Mediatek Inc., Hsinchu, Taiwan}
\mlsysaffiliation{ucr}{Department of Electrical and Computer Engineering, University of California at Riverside, Riverside, California, USA}
\mlsysaffiliation{princeton}{Electrical and Computer Engineering, Princeton University, New Jersey, USA}
\mlsysaffiliation{hkust}{Department of Computer Science and Engineering, The Hong Kong University of Science and Technology, Hong Kong SAR, China}
\mlsysaffiliation{hkust2}{Department of Computer Science and Engineering, The Hong Kong University of Science and Technology, Hong Kong SAR, China}

\mlsyscorrespondingauthor{Jinhyun So}{jinhyuns@usc.edu}
\mlsyscorrespondingauthor{Chaoyang He}{chaoyang.he@usc.edu}
\mlsyscorrespondingauthor{Salman Avestimehr}{avestimehr@ee.usc.edu}


\vskip 0.3in

\begin{abstract}
Secure model aggregation is a key component of federated learning (FL) that aims at protecting the privacy of each user’s individual model while allowing for their global aggregation.
It can be applied to any aggregation-based FL approach for training a global or personalized model.
Model aggregation needs to also be resilient against likely user dropouts in FL systems, making its design substantially more complex.
State-of-the-art secure aggregation protocols rely on secret sharing of the random-seeds used for mask generations at the users to enable the reconstruction and cancellation of those belonging to the dropped users.
The complexity of such approaches, however, grows substantially with the number of dropped users.
We propose a new approach, named \scheme, to overcome this bottleneck by changing the design from ``random-seed reconstruction of the dropped users'' to ``one-shot aggregate-mask reconstruction of the active users via mask encoding/decoding''.
%
%
We show that \scheme achieves the same privacy and dropout-resiliency guarantees as the state-of-the-art protocols while significantly reducing the overhead for resiliency against dropped users.
We also demonstrate that, unlike existing schemes, \scheme can be applied to secure aggregation in the \emph{asynchronous} FL setting.
Furthermore, we provide a modular system design and optimized on-device parallelization for scalable implementation, by enabling computational overlapping between model training and on-device encoding, as well as improving the speed of concurrent receiving and sending of chunked masks.
We evaluate \scheme via extensive experiments for training diverse models (logistic regression, shallow CNNs, MobileNetV3, and EfficientNet-B0) on various  datasets (MNIST, FEMNIST, CIFAR-10, GLD-23K) in a realistic FL system with large number of users and demonstrate that \scheme significantly reduces the total training time.
\end{abstract}
]

\printAffiliationsAndNotice{\mlsysEqualContribution}

\section{Introduction}\label{sec:intro}
Federated learning (FL) has emerged as a promising approach to enable distributed training  over a large number of users while protecting the privacy of each user~\cite{mcmahan2017communication,mcmahan2021advances,wang2021field}. The key idea of FL is to keep users' data on their devices and instead train local models at each user. The locally trained models are then aggregated via a server to update a global model, which is then pushed back to users. Due to model inversion attacks (e.g., ~\cite{NEURIPS2020_c4ede56b,wang2019beyond,zhu2020deep}), a critical consideration in FL design is to also ensure that the server does not learn the locally trained model of each user during  model aggregation. Furthermore, model aggregation should be robust against likely user dropouts (due to poor connectivity, low battery, unavailability, etc) in FL systems. As such, there have been a series of works that aim at developing secure aggregation protocols  for FL that protect the privacy of each user’s individual model while allowing their global aggregation amidst possible user dropouts \cite{bonawitz2017practical,kadhe2020fastsecagg,so2021turbo}. 

\begin{figure*}[t]
    \centering
    \includegraphics[width = 0.73\textwidth]{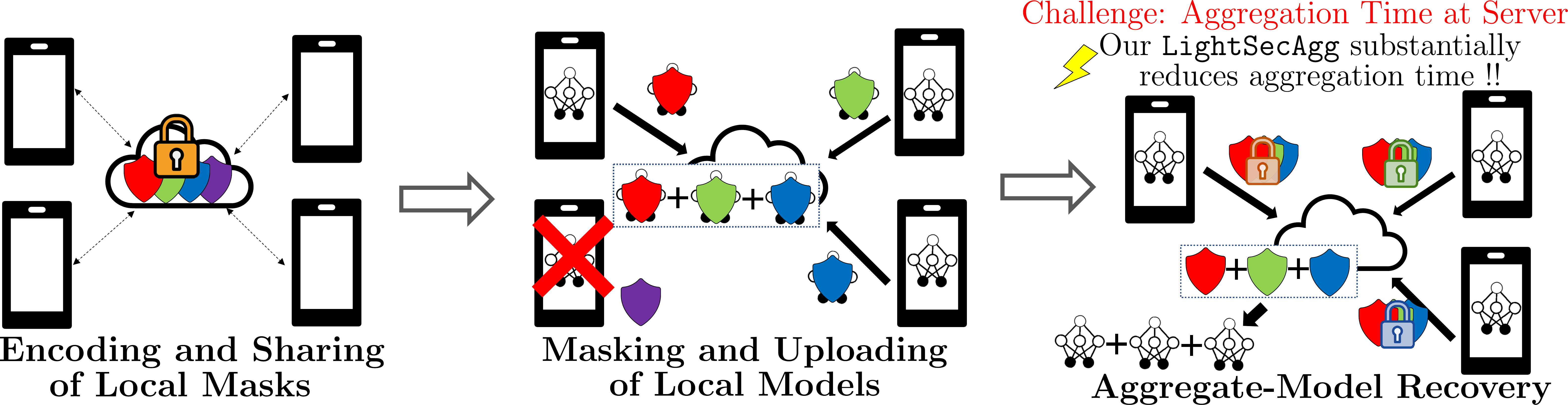}
    \vspace{-5pt}
    \caption{Illustration of our proposed \scheme protocol. (1) \textbf{Sharing encoded mask}: users encode and share their generated local masks. (2) \textbf{Masking model}: each user masks its model by random masks, and uploads its masked model to the server. (3) \textbf{Reconstructing aggregate-mask}: The surviving users upload the aggregate of encoded masks to reconstruct the desired aggregate-mask. The server recovers the aggregate-model by canceling out the reconstructed aggregate-mask.  } 
    \label{fig:intro}
    
\vspace{-7pt}
\end{figure*}

The state-of-the-art secure aggregation protocols essentially rely on two main principles: (1) pairwise random-seed agreement between users to generate masks that hide users’ models while having an additive structure that allows their cancellation when added at the server and (2) secret sharing of the random-seeds to enable the reconstruction and cancellation of masks belonging to dropped users. The main drawback of such approaches is that the number of mask reconstructions at the server substantially grows as more users are dropped, causing a major computational bottleneck. For instance, 
the execution time of the \google protocol proposed in ~\cite{bonawitz2017practical} is observed to be significantly limited by mask reconstructions at the server~\cite{MLSYS2019_bd686fd6}. \googlep~\cite{bell2020secure}, an improved version of \google, reduces the overhead at the server by replacing the complete communication graph of \google with a sparse random graph, such that secret sharing is only needed within a subset of users rather than all users pairs. However, the number of mask reconstructions in \googlep still increases as more users drop, 
eventually limits the scalability of FL systems. There have also been several other approaches, such as~\cite{so2021turbo,kadhe2020fastsecagg}, to alleviate this bottleneck, however they either increase round/communication complexity or compromise the dropout and privacy guarantees.

\textbf{Contributions.} We propose a new perspective for secure model aggregation in FL by turning the design focus from ``pairwise random-seed reconstruction of the dropped users’’ to ``one-shot aggregate-mask reconstruction of the surviving users’’. Using this viewpoint, we develop a new protocol named \scheme that provides the same level of privacy and dropout-resiliency guarantees as the state-of-the-art while substantially reducing the aggregation (hence runtime) complexity. As illustrated in Figure ~\ref{fig:intro}, the main idea of \scheme is that each user protects its local model using a locally generated random mask. This mask is then encoded and shared to other users in such a way that the aggregate-mask of any sufficiently large set of surviving users can be directly reconstructed at the server. In sharp contrast to prior schemes, in this approach the server only needs to reconstruct \emph{one} mask in the recovery phase, independent of the number of dropped users. 

Moreover, we provide a modular federated training system design and optimize on-device parallelization to improve the efficiency when secure aggregation and model training interact at the edge devices. This enables computational overlapping between model training and on-device encoding, as well as improving the speed of concurrent receiving and sending of chunked masks. To the best of our knowledge, this provides the first open-sourced and secure aggregation-enabled FL system that is built on the modern deep learning framework (PyTorch) and neural architecture (e.g., ResNet) with system and security co-design.

We further propose system-level optimization methods to improve the run-time. In particular, we design a federated training system and take advantage of the fact that the generation of random masks is independent of the computation of the local model, hence each user can parallelize these two operations via a multi-thread processing, which is beneficial to all evaluated secure aggregation protocols in reducing the total running time.

In addition to the synchronous FL setting, where all users train local models based on the same global model and the server performs a synchronized aggregation at each round, we also demonstrate that \scheme enables secure aggregation when no synchrony between users' local updates are imposed. This is unlike prior secure aggregation protocols, such as \google and \googlep, that are not compatible with  asynchronous FL. To the best of our knowledge, in the asynchronous FL setting, this is the first work to protect the privacy of the individual updates without relying on differential privacy \cite{truex2020ldp} or trusted execution environments (TEEs) \cite{nguyen2021federated}.

We run extensive experiments to empirically demonstrate the performance of \scheme in a real-world cross-device FL setting with up to $200$ users and compare it with two state-of-the-art protocols \google and \googlep.
To provide a comprehensive coverage of realistic FL settings, we train various machine learning models including logistic regression, convolutional neural network (CNN)~\cite{mcmahan2017communication}, MobileNetV3~\cite{howard2019searching}, and EfficientNet-B0~\cite{tan2019efficientnet}, for image classification over datasets of
different image sizes: low resolution images (FEMNIST~\cite{caldas2018leaf}, CIFAR-10~\cite{krizhevsky2009learning}), and high resolution images (Google Landmarks Dataset 23k~\cite{weyand2020google}). The empirical results show that \scheme provides significant speedup for all considered FL training tasks, achieving a performance gain of $8.5\times$-$12.7\times$ over \google and $2.9\times$-$4.4\times$ over \googlep, in realistic bandwidth settings at the users.
 Hence, compared to baselines, \scheme can even survive and speedup the training of large deep neural network models on high resolution image datasets. Breakdowns of the total running time further confirm that the primary gain lies in the complexity reduction at the server provided by \scheme, especially when the number of users are large.

{\bf Related works.}  Beyond the secure aggregation protocols proposed in ~\cite{bonawitz2017practical,bell2020secure}, there have been also other works that aim towards making secure aggregation more efficient.
\turbo~\cite{so2021turbo} utilizes a circular communication topology to reduce the communication overhead, but it incurs an additional round complexity and provides a weaker privacy guarantee than \google as it guarantees model privacy in the average sense rather than in the worst-case scenario. \fast~\cite{kadhe2020fastsecagg} reduces per-user overhead by using the Fast Fourier Transform multi-secret sharing, but it provides lower privacy and dropout guarantees compared to the other state-of-the-art protocols. The idea of one-shot reconstruction of the aggregate-mask was also employed in ~\cite{zhao2021information}, where the aggregated masks corresponding to each user dropout pattern was prepared by a trusted third party, encoded and distributed to users prior to model aggregation. 
The major advantages of \scheme over the scheme in ~\cite{zhao2021information} are 1) not requiring a trusted third party; and 2) requiring significantly less randomness generation and a much smaller storage cost at each user. 
Furthermore, there is also a lack of system-level performance evaluations of ~\cite{zhao2021information} in FL experiments. Finally, we emphasize that our \scheme protocol can be applied to any aggregation-based FL approach (e.g., \texttt{FedNova}\xspace~\cite{wang2020tackling}, \texttt{FedProx}\xspace~\cite{li2018federated}, \texttt{FedOpt}\xspace~\cite{asad2020fedopt}), personalized FL frameworks~\cite{NEURIPS2020_f4f1f13c,li2020ditto,fallah2020personalized,mushtaq2021spider,he2021ssfl}, communication-efficient FL~\cite{shlezinger2020uveqfed, reisizadeh2020fedpaq, elkordy2020secure}, and asynchronous FL, and their applications in computer vision \cite{he2021fedcv,He2020GroupKT,he2020fednas}, natural language processing \cite{lin2021fednlp,he2021pipetransformer}, data mining \cite{he2021fedgraphnn,ezzeldin2021fairfed,liang2021omnilytics,he2021spreadgnn,he2019central}, and Internet of things (IoTs) \cite{Zhang2021FederatedLF,Zhang2021FederatedLF2}. 

\section{Problem Setting}\label{sec:sys}

FL is a distributed training framework for machine learning
, where the goal is to learn a global model $\mathbf{x}$ with dimension $d$ using data held at edge devices. This can be represented by minimizing a global objective function $F$: $F(\mathbf{x}) = \sum^N_{i=1}p_iF_i(\mathbf{x})$, where $N$ is the total number of users, $F_i$ is the local objective function of user $i$, and $p_i\geq 0$ is a weight parameter assigned to user $i$ to specify the relative impact of each user such that $\sum^N_{i=1}p_i=1$.\footnote{For simplicity, we assume that all users have equal-sized datasets, i.e., $p_i = \frac{1}{N}$ for all $i \in [N]$.} 

Training in FL is performed through an iterative process, where the  users interact through a  server to update the global model. At each iteration, the server shares the current global model, denoted by $\mathbf{x}(t)$, with the edge users. Each user $i$ creates a local update, $\mathbf{x}_i(t)$. The local models are sent to the server and then aggregated by the server. Using the aggregated models, the server updates the global model $\mathbf{x}(t+1)$ for the next iteration. In FL, some users may potentially drop from the learning procedure for various reasons such as having unreliable communication connections. The goal of the server is to obtain the sum of the surviving users' local models. This update equation is given by $\mathbf{x}(t+1) =  \frac{1}{|{\cal U}(t)|}\sum_{i \in \mathcal{U}(t)}\mathbf{x}_i(t)$, where $\mathcal{U}(t)$ denotes the set of surviving users at iteration $t$. Then, the server pushes the updated global model $\mathbf{x}(t+1)$ to the edge users.


Local models carry extensive information about the users' datasets, and in fact their private data  can be reconstructed from the local models by using a model inversion attack~\cite{NEURIPS2020_c4ede56b,wang2019beyond,zhu2020deep}. To address this privacy leakage from local models, secure aggregation has been introduced in~\cite{bonawitz2017practical}. A secure aggregation protocol enables the computation of the aggregated global model while ensuring that the server (and other users) learn no information about the local models beyond their aggregated model. In particular, the goal is to securely recover the aggregate of the local models $ \mathbf{y} = \sum_{i \in \mathcal{U}}\mathbf{x}_i$, where the iteration index $t$ is omitted for simplicity. Since secure aggregation protocols build on cryptographic primitives that require all operations to be carried out over a finite field, we assume that the elements of $\mathbf{x}_i$ and $\mathbf{y}$ are from a finite field $\mathbb{F}_q$
for some field size $q$. We require a secure aggregation protocol for FL to have the following key features.
\begin{itemize}[leftmargin=*, itemsep=1pt]
    \item {\bf Threat model and privacy guarantee.} We consider a threat model where the users and the server are honest but curious. We assume that up to $T$ (out of $N$) users can collude with each other as well as with the server to infer the local models of other users. The secure aggregation protocol has to guarantee that nothing can be learned beyond the aggregate-model, even if up to $T$ users cooperate with each other. We consider privacy leakage in the strong information-theoretic sense. This requires that for every subset of users $\mathcal{T} \subseteq [N]$ of size at most $T$, we must have mutual information $I(\{\mathbf{x}_i\}_{i \in [N]};\mathbf{Y}|\sum_{i \in \mathcal{U}}\mathbf{x}_i, \mathbf{Z}_{\mathcal{T}}) = 0$, where $\mathbf{Y}$ is the collection of information 
   at the server, and $\mathbf{Z}_{\mathcal{T}}$ is the collection of information at the users in $\mathcal{T}$.

    \item {\bf Dropout-resiliency guarantee.} In the FL setting, it is common for users to be dropped or delayed at any time during protocol execution for various reasons, e.g., delayed/interrupted processing, poor wireless channel conditions, low battery, etc. We assume that there are at most $D$ dropped users during the execution of protocol, i.e., there are at least $N-D$ surviving users after potential dropouts. The protocol must guarantee that the server can correctly recover the aggregated models of the surviving users, even if up to $D$ users drop. 

    \item {\bf Applicability to asynchronous FL.} 
    Synchronizing all users for training at each round of FL can be slow and costly, especially when the number of users are large. Asynchronous FL handles this challenge by incorporating the updates of the users in asynchronous fashion \cite{xie2019asynchronous,van2020asynchronous,chai2020fedat,chen2020asynchronous}.
    This asynchrony, however, creates a mismatch of staleness among the users, which causes the incompatibility of the existing secure aggregation protocols (such as \cite{bonawitz2017practical, bell2020secure}). More specifically, since it is not known a priori which local models will be aggregated together, the current secure aggregation protocols that are based on pairwise random masking among the users fail to work.
    We aim at designing a versatile secure aggregation protocol that is applicable to both synchronous and asynchronous FL.
\end{itemize} 
 \textbf{Goal.} We aim to design an efficient and scalable secure aggregation protocol that simultaneously achieves strong privacy and dropout-resiliency guarantees, scaling linearly with the number of users $N$, e.g., simultaneously achieves privacy guarantee $T =\frac{N}{2}$ and dropout-resiliency guarantee $D = \frac{N}{2}-1$. Moreover, the protocol should be compatible with both synchronous and asynchronous FL.

\section{Overview of Baseline Protocols: \google and \googlep}\label{sec:overviews}

We first review the state-of-the-art secure aggregation protocols  \google~\cite{bonawitz2017practical} and \googlep~\cite{bell2020secure} as our baselines. 
Essentially, \google and \googlep require each user to mask its local model using random keys before aggregation. In \google, the privacy of the individual models is protected by pairwise random masking. Through a key agreement (e.g., Diffie-Hellman~\cite{diffie1976new}), each pair of users $i,j \in [N]$ agree on a pairwise random seed $a_{i,j} = \text{Key.Agree}(sk_i,pk_j) = \text{Key.Agree}(sk_j,pk_i)$ where $sk_i$ and $pk_i$ are the private and public keys of user $i$, respectively. In addition, user $i$ creates a private random seed $b_i$ to prevent the privacy breaches that may occur if user $i$ is only delayed rather than dropped, in which case the pairwise masks alone are not sufficient for privacy protection. User $i \in [N]$ then masks its model $\mathbf{x}_i$ as $\mathbf{\tilde{x}}_i = \mathbf{x}_i + \text{PRG}(b_i) + \sum_{j:i<j}\text{PRG}(a_{i,j}) - \sum_{j:i>j}\text{PRG}(a_{j,i})$, where PRG is a pseudo random generator, and sends it to the server. Finally, user $i$ secret shares its private seed $b_i$ as well as private key $sk_i$ with the other users via Shamir’s secret sharing~\cite{yao1982protocols}. From the subset of users who survived the previous stage, the server collects either the shares of the private key belonging to a dropped user, or the shares of the private seed belonging to a surviving user (but not both). Using the collected shares, the server reconstructs the private seed of each surviving user, and the pairwise seeds of each dropped user. The server then computes the aggregated model as follows
\begin{align}
\sum_{i \in \mathcal{U}}\mathbf{x}_i &= \sum_{i \in \mathcal{U}} (\mathbf{\tilde{x}}_i - \text{PRG}(b_i)) \notag\\ 
&+ \sum_{i \in \mathcal{D}}\left(  \sum_{j:i<j}\text{PRG}(a_{i,j}) - \sum_{j:i>j}\text{PRG}(a_{j,i})\right),
\end{align}
where $\mathcal{U}$ and $\mathcal{D}$ represent the set of surviving and dropped users, respectively. \google protects model privacy against $T$ colluding users and is robust to $D$ user dropouts as long as $N-D>T$.

\begin{figure}[t!]
    \centering
    \includegraphics[width = 0.48\textwidth]{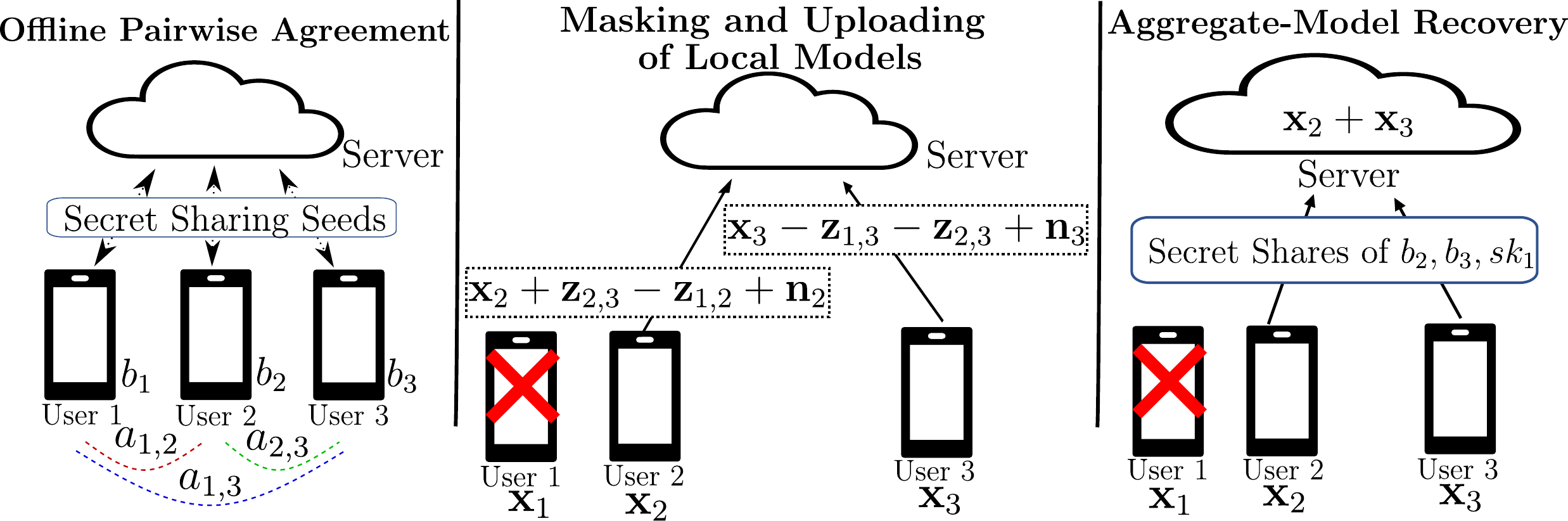}\vspace{-2mm}
    \caption{An illustration of \google in the example of $3$ users is depicted. The users first agree on pairwise random seeds, and secret share their private random seeds and private keys. The local models are protected by the pairwise random masking. Suppose that user $1$ drops. To recover the aggregate-mask, the server first reconstructs the private random seeds of the surviving users and the private key of user $1$ by collecting the secret shares for each of them. Then, the server recovers $\mathbf{z}_{1,2}$, $\mathbf{z}_{1,3}$, $\mathbf{n}_2$ and $\mathbf{n}_3$, which incurs the computational cost of $4d$ at the server.
   } 
    \label{fig:SecAgg}
    \vspace{-3mm}
\end{figure}

We now illustrate \google through a simple example. Consider a secure aggregation problem in FL, where there are $N = 3$ users with $T=1$ privacy guarantee and dropout-resiliency guarantee $D=1$. Each user $i \in \{1,2,3\}$ holds a local model $\mathbf{x}_i \in \mathbb{F}^d_q$ where $d$ is the model size and $q$ is the size of the finite field. As shown in Figure \ref{fig:SecAgg}, \google is composed of the following three phases.

{\bf Offline pairwise agreement.} User $1$ and user $2$ agree on pairwise random seed $a_{1,2}$. User $1$ and user $3$ agree on pairwise random seed $a_{1,3}$. User $2$ and user $3$ agree on pairwise random seed $a_{2,3}$. In addition, user $i \in \{1,2,3\}$ creates a private random seed $b_i$. Then, user $i$ secret shares $b_i$ and its private key $sk_i$ with the other users via Shamir’s secret sharing. In this example, a $2$ out of $3$ secret sharing is used to tolerate $1$ curious user.

{\bf Masking and uploading of local models.} To provide the privacy of each individual model, user $i \in \{1,2,3\}$ masks its model $\mathbf{x}_i$ as follows:
\begin{align*}
&\mathbf{\tilde{x}}_1 = \mathbf{x}_1 + \mathbf{n}_1 + \mathbf{z}_{1,2} + \mathbf{z}_{1,3}, \notag \\ &\mathbf{\tilde{x}}_2 = \mathbf{x}_2 + \mathbf{n}_2 + \mathbf{z}_{2,3} - \mathbf{z}_{1,2}, \notag \\ &\mathbf{\tilde{x}}_3 = \mathbf{x}_3 + \mathbf{n}_3 - \mathbf{z}_{1,3} - \mathbf{z}_{2,3}, \notag
\end{align*}
where $\mathbf{n}_i = \text{PRG}(b_i)$ and $\mathbf{z}_{i,j} = \text{PRG}(a_{i,j})$ are the random masks generated by a pseudo random generator. Then user $i \in \{1,2,3\}$ sends its masked local model $\mathbf{\tilde{x}}_i$ to the server.  

{\bf Aggregate-model recovery.} Suppose that user $1$ drops in the previous phase. The goal of the server is to compute the aggregate of models $\mathbf{x}_2+\mathbf{x}_3$. Note that
\begin{align}
\label{eq:secAggEx}
 \mathbf{x}_2+ \mathbf{x}_3 = \mathbf{\tilde{x}}_2+\mathbf{\tilde{x}}_3+(\mathbf{z}_{1,2}+\mathbf{z}_{1,3}-\mathbf{n}_2 - \mathbf{n}_3).
\end{align}
Hence, the server needs to reconstruct masks $\mathbf{n}_2$, $\mathbf{n}_3$, $\mathbf{z}_{1,2}$, $\mathbf{z}_{1,3}$ to recover $\mathbf{x}_2+\mathbf{x}_3$. To do so, the server has to collect two shares for each of $b_2$, $b_3$, $sk_1$, and then compute the aggregate model by (\ref{eq:secAggEx}). Since the complexity of evaluating a PRG scales linearly with its size, the computational cost of the server for mask reconstruction is $4d$.

We note that \google requires the server to compute a PRG function on \emph{each} of the reconstructed seeds to recover the aggregated masks, which incurs the overhead
of $O(N^2)$ (see more details in Section~\ref{sec:analysis}) and dominates the overall execution time of the protocol~\cite{bonawitz2017practical,MLSYS2019_bd686fd6}.
\googlep reduces the overhead of mask reconstructions from $O(N^2)$ to $O(N\log{N})$ by replacing the complete communication graph of \google with a sparse random graph of degree $O(\log{N})$ to reduce both communication and computational loads. Reconstructing pairwise random masks in \google and \googlep poses a major bottleneck in scaling to a large number of users.  

\begin{remark}\label{rmk:Incompatibility}(Incompatibility of \google and \googlep with Asynchronous FL). \normalfont
It is important to note that \google and \googlep cannot be applied to asynchronous FL as the cancellation of the pairwise random masks based on the key agreement protocol is not guaranteed. This is because the users do not know a priori which local models will be aggregated together, hence the masks cannot be designed to cancel out in these protocols. We explain this in more detail in Appendix \ref{app:sub:incompatibility}. 
It is also worth noting that a recently proposed protocol known as \FedBuff \cite{nguyen2021federated} enables secure aggregation in asynchronous FL through a trusted execution environment (TEE)-enabled buffer, where the server stores the local models that it receives in this \emph{private} buffer. The reliance of \FedBuff on TEEs, however, limits the buffer size in this approach as TEEs have limited memory. It would also limit its application to FL settings where TEEs are available.
\end{remark}
\color{black}
\vspace{-10pt}

\section{\scheme Protocol}\label{sec:scheme}

\begin{figure}[!t]
    \centering
    \includegraphics[width = 0.48\textwidth]{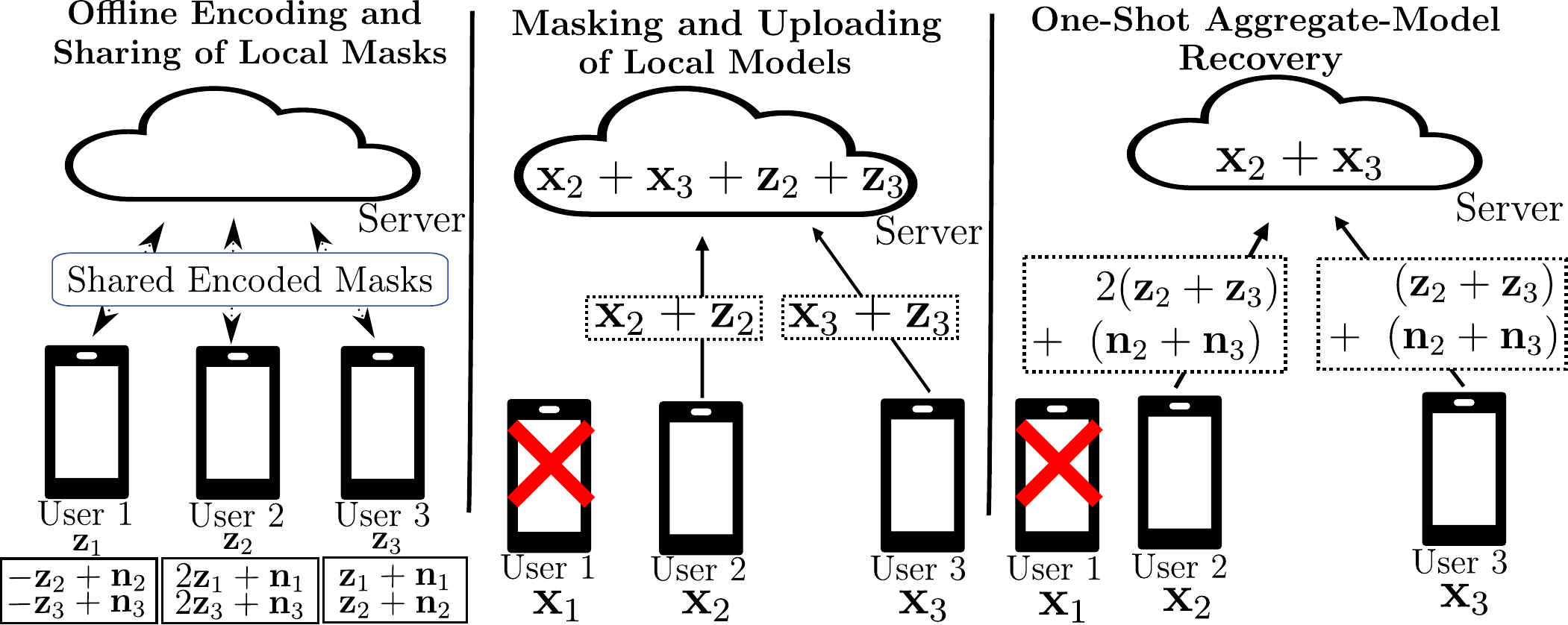}\vspace{-2mm}
    \caption{An illustration of \scheme in the example of $3$ users is depicted. Each user first generates a single mask. Each mask of a user is encoded and shared to other users. Each user's local model is protected by its generated mask. Suppose that user $1$ drops during the execution of protocol. The server directly recovers the aggregate-mask in one shot. In this example, \scheme reduces the computational cost at the server from $4d$ to $d$.  } 
    \label{fig:LightSecAgg}
    \vspace{-10pt}
\end{figure}

Before providing a general description of \scheme, we first illustrate its key ideas through the previous 3-user example in the synchronous setting.
As shown in Figure \ref{fig:LightSecAgg}, \scheme has the following three phases.

{\bf Offline encoding and sharing of local masks.} User $i \in \{1,2,3\}$ randomly picks $\mathbf{z}_i$ and $\mathbf{n}_i$ from $\mathbb{F}^d_q$. User $i \in \{1,2,3\}$ creates the masked version of $\mathbf{z}_i$ as 
\begin{align*}
&\mathbf{\tilde{z}}_{1,1} = - \mathbf{z}_1 + \mathbf{n}_1, \ \mathbf{\tilde{z}}_{1,2} = 2\mathbf{z}_1 + \mathbf{n}_1, \ \mathbf{\tilde{z}}_{1,3} = \mathbf{z}_1 + \mathbf{n}_1; \notag \\
&\mathbf{\tilde{z}}_{2,1} = - \mathbf{z}_2 + \mathbf{n}_2, \ \mathbf{\tilde{z}}_{2,2} = 2\mathbf{z}_2 + \mathbf{n}_2, \ \mathbf{\tilde{z}}_{2,3} = \mathbf{z}_2 + \mathbf{n}_2; \notag \\ 
&\mathbf{\tilde{z}}_{3,1} = - \mathbf{z}_3 + \mathbf{n}_3, \ \mathbf{\tilde{z}}_{3,2} = 2\mathbf{z}_3 + \mathbf{n}_3, \ \mathbf{\tilde{z}}_{3,3} = \mathbf{z}_3 + \mathbf{n}_3; \notag
\end{align*}
and user $i\in \{1,2,3\}$ sends $\mathbf{\tilde{z}}_{i,j}$ to each user $j \in \{1,2,3\}$. Thus, user $i\in \{1,2,3\}$ receives $\tilde{\mathbf{z}}_{j,i}$ for $j \in \{1,2,3\}$. In this case, this procedure provides robustness against $1$ dropped user and privacy against $1$ curious user.

{\bf Masking and uploading of local models.} To make each individual model private, each user $i \in \{1,2,3\}$ masks its local model as follows:
\begin{align}
    \mathbf{\tilde{x}}_1 = \mathbf{x}_1 + \mathbf{z}_1, \quad    \mathbf{\tilde{x}}_2 = \mathbf{x}_2 + \mathbf{z}_2,\quad
     \mathbf{\tilde{x}}_3 = \mathbf{x}_3 + \mathbf{z}_3,
\end{align}
and sends its masked model to the server. 

{\bf One-shot aggregate-model recovery.} Suppose that user $1$ drops in the previous phase. To recover the aggregate of models $\mathbf{x}_2 + \mathbf{x}_3$, the server only needs to know the aggregated masks $\mathbf{z}_2+\mathbf{z}_3$. To recover $\mathbf{z}_2+\mathbf{z}_3$, the surviving user $2$ and user $3$ send $\tilde{\mathbf{z}}_{2,2} + \tilde{\mathbf{z}}_{3,2}$ and $\mathbf{\tilde{z}}_{2,3} + \mathbf{\tilde{z}}_{3,3}$,
\begin{align}
    &\mathbf{\tilde{z}}_{2,2} + \mathbf{\tilde{z}}_{3,2} = 2(\mathbf{z}_2+\mathbf{z}_3)+\mathbf{n}_2+\mathbf{n}_3, \notag \\
    &\mathbf{\tilde{z}}_{2,3} + \mathbf{\tilde{z}}_{3,3} = (\mathbf{z}_2+\mathbf{z}_3)+\mathbf{n}_2+\mathbf{n}_3, \notag
\end{align}
to the server, respectively. After receiving the messages from user $2$ and user $3$, the server can directly recover the aggregated masks via an one-shot computation as follows:
\begin{align}
    \mathbf{z}_2+\mathbf{z}_3 = \mathbf{\tilde{z}}_{2,2} + \mathbf{\tilde{z}}_{3,2} - (\mathbf{\tilde{z}}_{2,3} + \mathbf{\tilde{z}}_{3,3}).
\end{align}
Then, the server recovers the aggregate-model $\mathbf{x}_2 + \mathbf{x}_3$ by subtracting $\mathbf{z}_2+\mathbf{z}_3$ from $\mathbf{\tilde{x}}_2 + \mathbf{\tilde{x}}_3$. As opposed to \google which has to reconstruct the random seeds of the dropped users, \scheme enables the server to reconstruct the desired aggregate of masks via a one-shot recovery. Compared with \google, \scheme reduces the server's computational cost from $4d$ to $d$ in this simple example.

\subsection{General Description of \scheme for Synchronous FL}
We formally present \scheme, whose idea is to encode the local generated random masks in a way that the server can recover the aggregate of masks from the encoded masks via an one-shot computation with a cost that does not scale with $N$. \scheme has three design parameters: (1) $0 \leq T \leq N-1$ representing the privacy guarantee; (2) $0\leq D \leq N-1$ representing the dropout-resiliency guarantee; (3) $1\leq U \leq N$ representing the targeted number of surviving users. In particular, parameters $T$, $D$, and $U$ are selected such that $N-D\geq U > T \geq 0$.
 
\scheme is composed of three main phases. First, each user first partitions its local random mask to $U-T$ pieces and creates encoded masks via a Maximum Distance Separable (MDS) code \cite{roth1989mds,yu2019lagrange,tang2021verifiable,so2021codedprivateml} to provide robustness against $D$ dropped users and privacy against $T$ colluding users. Each user sends one of the encoded masks to one of the other users for the purpose of one-shot recovery. Second, each user uploads its masked local model to the server. Third,  the server reconstructs the aggregated masks of the surviving users to recover their aggregate of models. Each surviving user sends the aggregated encoded masks to the server. After receiving $U$ aggregated encoded masks from the surviving users, the server recovers the aggregate-mask and the desired aggregate-model. The pseudo code of \scheme is provided in Appendix~\ref{appendix:code}. We now describe each of these phases in detail.

\textbf{Offline encoding and sharing of local masks.}
User $i \in [N]$ picks $\mathbf{z}_i$ uniformly at random from $\mathbb{F}^d_q$ and partitions it to $U-T$ sub-masks $[\mathbf{z}_i]_k \in \mathbb{F}^{\frac{d}{U-T}}_q$, $k \in [U-T]$. With the randomly picked $[\mathbf{n}_i]_k\in \mathbb{F}^{\frac{d}{U-T}}_q$ for $k \in \{U-T+1,\dots,U\}$, user $i \in [N]$ encodes sub-masks $[\mathbf{z}_i]_k$'s as
\begin{align} \label{eq:def_z_tilde} 
    [\mathbf{\tilde{z}}_i]_j =([\mathbf{z}_i]_1,\dots, [\mathbf{z}_i]_{U-T},[\mathbf{n}_i]_{U-T+1},\dots ,[\mathbf{n}_i]_U)\cdot W_j,
\end{align}
where $W_j$ is $j$'th column of a \emph{$T$-private} MDS matrix $W \in \mathbb{F}^{U\times N}_q$.
In particular, we say an MDS matrix\footnote{A matrix $W \in \mathbb{F}^{U\times N}_q$ ($U<N$) is an MDS matrix if any $U\times U$ sub-matrix of $W$ is non-singular.} is {$T$-private} {iff} the submatrix consisting of its $\{U-T+1,...,U\}$-th rows is also MDS. 
A $T$-private MDS matrix guarantees that $I(\mathbf{z}_i;\{[\mathbf{\tilde{z}}_i]_j\}_{j \in \mathcal{T}})= 0$, for any $i\in[N]$ and any $\mathcal{T \subseteq[N]}$ of size $T$, if $[\mathbf{n}_i]_k$'s are jointly uniformly random.
We can always find $T$-private MDS matrices for any $U$, $N$, and $T$ (e.g.,~\cite{shamir1979share,yu2019lagrange,roth1989mds}). 
Each user $i\in [N]$ sends $[\mathbf{\tilde{z}}_i]_j$ to user $j \in [N]\backslash \{i\}$. In the end of offline encoding and sharing of local masks, each user $i\in [N]$ has $[\mathbf{\tilde{z}}_j]_i$ from $j \in [N]$.\footnote{All users communicate through secure (private and authenticated) channels, i.e., the server would only receive the encrypted version of $[\mathbf{\tilde{z}}_i]_{j}$'s. Such secure communication is also used in prior works on secure aggregation, e.g., \google, \googlep.}

\textbf{Masking and uploading of local models.} To protect the local models, each user $i$ masks its local model as $\mathbf{\tilde{x}}_i = \mathbf{x}_i + \mathbf{z}_i$, and sends it to the server. Since some users may drop in this phase, the server identifies the set of surviving users, denoted by $\mathcal{U}_1 \subseteq [N]$. The server intends to recover $\sum_{i \in \mathcal{U}_1}\mathbf{x}_i$.
We note that before masking the model, each user quantizes the local model to convert from the domain of real numbers to the finite field (Appendix \ref{subsubsec:BASecAgg_secondphase}).

\textbf{One-shot aggregate-model recovery.} After identifying the surviving users in the previous phase, user $j\in \mathcal{U}_1$ is notified to send its aggregated encoded sub-masks $\sum_{i\in \mathcal{U}_1}[\mathbf{\tilde{z}}_i]_j$ to the server for the purpose of one-shot recovery. We note that each $\sum_{i\in \mathcal{U}_1}[\mathbf{\tilde{z}}_i]_j$ is an encoded version of $\sum_{i \in \mathcal{U}_1}[\mathbf{z}_i]_k$ for $k \in [U-T]$ using the MDS matrix $W$ (see more details in Appendix~\ref{appendix:main_thm}). 
Thus, the server is able to recover $\sum_{i \in \mathcal{U}_1}[\mathbf{z}_i]_k$ for $k \in [U-T]$ via MDS decoding after receiving a set of any $U$ messages from the participating users. 
The server obtains the aggregated masks $\sum_{i \in \mathcal{U}_1}\mathbf{z}_i$ by concatenating $\sum_{i \in \mathcal{U}_1}[\mathbf{z}_i]_k$'s. Lastly, the server recovers the desired aggregate of models for the set of participating users $\mathcal{U}_1$ by subtracting $\sum_{i \in \mathcal{U}_1}\mathbf{z}_i$ from $\sum_{i \in \mathcal{U}_1}\mathbf{\tilde{x}}_i$.



\begin{remark}\normalfont
Note that it is not necessary to have a stable communication link between every pair of users in \scheme. Specifically, given the design parameter $U$, \scheme only requires at least $U$ surviving users at any time during the execution. That is, even if up to $N-U$ users drop or get delayed due to unstable communication links, the server can still reconstruct the aggregate-mask.  
\end{remark}
\vspace{-3mm}
\begin{remark}\normalfont
We note that \scheme directly applies for secure aggregation of weighted local models. The sharing of the masking keys among the clients does not require the knowledge of the weight coefficients. For example, \scheme can work for the case in which all users do not have equal-sized datasets. Suppose that user $i$ holds a dataset with a number of samples $s_i$. Rather than directly masking the local model $\mathbf{x}_i$, user $i$ first computes $\mathbf{x}^{'}_i = s_i\mathbf{x}_i$. Then, user $i$ uploads $\mathbf{x}^{'}_i + \mathbf{z}_i$ to the server. 
Through the \scheme protocol, the server can recover $\sum_{i \in \mathcal{U}}\mathbf{x}^{'}_i = \sum_{i \in \mathcal{U}}s_i\mathbf{x}_i$ securely. By dividing by $\sum_{i \in \mathcal{U}}s_i$, the server can obtain the desired aggregate of weighted model $\sum_{i \in \mathcal{U}}p_i\mathbf{x}_i$ where $p_i = \frac{s_i}{\sum_{i \in \mathcal{U}}s_i}$.
\end{remark}

\subsection{Extension to Asynchronous FL}\label{subsec:ext_to_async}
We now describe how \scheme can be applied to asynchronous FL. We consider the asynchronous FL setting  with bounded staleness as considered in \cite{nguyen2021federated}, where the updates of the users are not synchronized and the staleness of each user is bounded by $\tau_{\mathrm{max}}$. In this setting, the server stores the models that it receives in a buffer of size $K$ and updates the global model once the buffer is full. More generally, \scheme may apply to any asynchronous FL setting where a group of local models are aggregated at each round. That is, the group size does not need to be fixed in all rounds. While the baselines are not compatible with this setting, \scheme can be applied by encoding the local masks in a way that the server can recover the aggregate of masks from the encoded masks via a one-shot computation, even though the masks are generated in different training rounds. 
Specifically, the users share the encoded masks with the timestamp to figure out which encoded masks should be aggregated for the reconstruction of the aggregate of masks. 
As the users aggregate the encoded masks after the server stores the local updates in the buffer, the users can aggregate the encoded masks according to the timestamp of the stored updates.
Due to the commutative property of MDS coding and addition, the server can reconstruct the aggregate of masks even though the masks are generated in different training rounds. 
We postpone the detailed description of the \scheme protocol for the asynchronous setting to Appendix \ref{appendix:asyncFL}.







\section{Theoretical Analysis}\label{sec:analysis}
\subsection{Theoretical Guarantees}

We now state our main result for the theoretical guarantees of the \scheme protocol.
\begin{theorem}\label{thm:main_thm}\normalfont
Consider a secure aggregation problem in federated learning with $N$ users. Then, the proposed \scheme protocol can \emph{simultaneously} achieve (1) privacy guarantee against up to any $T$ colluding users, and (2) dropout-resiliency guarantee against up to any $D$ dropped users, for any pair of privacy guarantee $T$ and dropout-resiliency guarantee $D$ such that $T+D<N$. 
\end{theorem}
The proof of Theorem \ref{thm:main_thm}, which is applicable to both synchronous and asynchronous FL settings, is presented in Appendix \ref{appendix:main_thm}.

\begin{remark}\normalfont
Theorem~\ref{thm:main_thm} provides a trade-off between privacy and dropout-resiliency guarantees, i.e., \scheme can increase the privacy guarantee by reducing the dropout-resiliency guarantee and vice versa. As \google~\cite{bonawitz2017practical}, \scheme achieves the worst-case dropout-resiliency guarantee. That is, for any privacy guarantee $T$ and the number of dropped users $D<N-T$, \scheme ensures that any set of dropped users of size $D$ in secure aggregation can be tolerated. Differently, \googlep~\cite{bell2020secure}, \fast~\cite{kadhe2020fastsecagg}, and \turbo~\cite{so2021turbo} relax the worst-case constraint to random dropouts and provide a probabilistic dropout-resiliency guarantee, i.e., the desired aggregate-model can be correctly recovered with high probability. 
\end{remark}

\begin{remark}\normalfont
From the training convergence perspective, \scheme only adds a quantization step to the local model updates of the users. The impact of this model quantization on FL convergence is well studied in the synchronous FL~\cite{reisizadeh2020fedpaq, elkordy2020secure}. In the asyncrhonous FL, however, we need to analyze the convergence of \scheme. We provide this analysis in the smooth and non-convex setting in Appendix \ref{app:sub:async-convergence}.
\end{remark}
\color{black}


\subsection{Complexity Analysis of \scheme}\label{sec:complexity}


We measure the storage cost, communication load, and computational load of \scheme in units of elements or operations in $\mathbb{F}_q$ for a single training round. 
Recall that $U$ is a design parameter chosen such that $N-D\geq U > T$.

\noindent {\bf Offline storage cost.} Each user $i$ independently generates a random mask $\mathbf{z}_i$ of length $d$. Additionally, each user $i$ stores a coded mask 
$[\mathbf{\tilde{z}}_j]_i$ of size $\frac{d}{U-T}$, for $j\in[N]$. Hence, the total offline storage cost at each user is $(1+\frac{N}{U-T})d$.

\noindent {\bf Offline communication and computation loads.} For each iteration of secure aggregation, before the local model is computed, each user prepares offline coded random masks and distributes them to the other users. Specifically, each user encodes $U$ local data segments with each of size $\frac{d}{U-T}$ into $N$ coded segments and distributes each of them to one of $N$ users. Hence, the offline computational and communication load of \scheme at each user is $O(\frac{dN \log N}{U-T})$ and $O(\frac{dN}{U-T})$, respectively. 

\noindent {\bf Communication load during aggregation.} While each user uploads a masked model of length $d$, in the phase of aggregate-model recovery, no matter how many users drop, each surviving user in $\mathcal{U}_1$ sends a coded mask of size $\frac{d}{U-T}$. The server is guaranteed to recover the aggregate-model of the surviving users in ${\cal U}_1$ after receiving messages from any $U$ users. The total required communication load at the server in the phase of mask recovery is therefore $\frac{U}{U-T}d$. 

\noindent {\bf Computation load during aggregation.} 
The major computational bottleneck of \scheme is the decoding process to recover $\sum_{j \in {\cal U}_1} \mathbf{z}_j$ at the server. This involves decoding a $U$-dimensional MDS code from $U$ coded symbols, which can be performed with $O(U\log U)$ operations on elements in $\mathbb{F}_q$, hence a total computational load of $\frac{U\log U}{U-T}d$.

\renewcommand{\arraystretch}{1.5}
\begin{table}[!t]
\centering
\caption{Complexity comparison between \google, \googlep, and \scheme. Here $N$ is the total number of users, $d$ is the model size, $s$ is the length of the secret keys as the seeds for PRG ($s \ll d$). In the table, U stands for User and S stands for Server.}
\vspace{1mm}
\footnotesize
\resizebox{\linewidth}{!}{
  \begin{tabular}{|c|c|c|c|}
  \hline 
& \google & \googlep & \scheme  \\
    \hline
offline comm. (U)  & $O(sN)$ & $O(s\log{N})$ & $O(d)$  \\ 
    \hline
offline comp. (U) & $O(dN + sN^2)$ & $O(d\log{N}+s\log^2 N)$ & $O(d\log{N})$  \\ 
    \hline
online comm. (U) & $O(d + sN)$& $O(d+s\log{N})$ &  $O(d)$ \\ 
    \hline
online comm. (S) & $O(dN + sN^2)$ & $O(dN+sN\log{N})$&  $O(dN)$ \\ 
    \hline
online comp. (U)& $O(d)$ & $O(d)$ & $O(d)$ \\ 
    \hline
reconstruction (S) & 
$O(dN^2)$
& 
$O(dN\log{N})$ 
& $O(d\log{N} )$  \\
    \hline
  \end{tabular}
 }
  \label{table:compare_google}
\vspace{-0.6cm}
\end{table}

We compare the communication and computational complexities of \scheme with baseline protocols. In particular, we consider the case where secure aggregation protocols aim at providing privacy guarantee $T = \frac{N}{2}$ and dropout-resiliency guarantee $D = pN$ simultaneously for some $0\leq p < \frac{1}{2}$. As shown in Table~\ref{table:compare_google}, by choosing $U = (1-p)N$, \scheme significantly improves the computational efficiency at the server during aggregation. \google and \googlep incurs a total computational load of $O(dN^2)$ and $O(d N\log{N})$, respectively at the server, while the server complexity of \scheme remains nearly constant with respect to $N$. It is expected to substantially reduce the overall aggregation time for a large number of users, which is bottlenecked by the server's computation in \google~\cite{bonawitz2017practical,MLSYS2019_bd686fd6}. More detailed discussions, as well as a comparison with another recently proposed secure aggregation protocol~\cite{zhao2021information}, which achieves similar server complexity as \scheme, are carried out in Appendix~\ref{appendix:discussion}.

\section{System Design and Optimization}\label{sec:system}
Apart from theoretical design and analysis, we have further designed a FL training system to reduce the overhead of secure model aggregation and enable realistic evaluation of \scheme in cross-device FL. 
\begin{figure}[h!]
    \centering
    \includegraphics[width=1.0\linewidth]{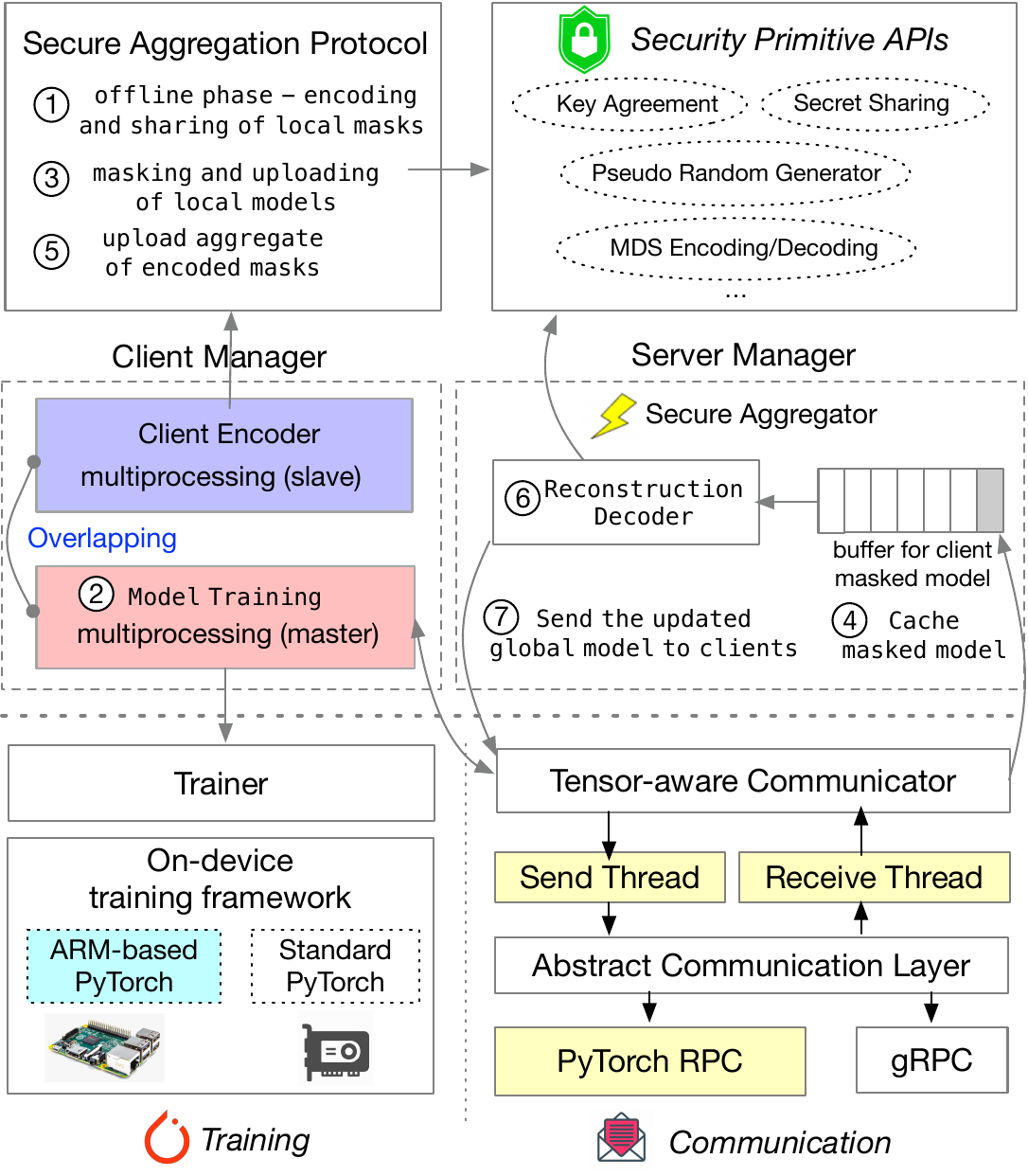}
    \vspace{-0.3cm}
    \caption{Overview of the System Design}
    \label{fig:overview_design}
    \vspace{-0.5cm}
\end{figure}

The software architecture is shown in Figure \ref{fig:overview_design}. In order to keep the software architecture lightweight and maintainable, we do not over-design and only modularize the system as the foundation layer and the algorithm layer. 

The foundation layer (blocks below the dashed line) contains the communicator and training engine. The communicator can support multiple communication protocols (PyTorch RPC \cite{trpc}, and gRPC \cite{grpc}), but it provides a unified communication interface for the algorithmic layer. In the training engine, in addition to standard PyTorch for GPU, we also compile the ARM-based PyTorch for embedded edge devices (e.g., Raspberry Pi). 

In the algorithm layer, \texttt{Client Manager} calls \texttt{Trainer} in the foundation layer to perform on-device training. \texttt{Client Manager} also integrates \texttt{Client Encoder} to complete the secure aggregation protocol, which is supported by security primitive APIs. In \texttt{Server Manager}, \texttt{Secure Aggregator} maintains the cache for masked models, and once the cache is full, it starts reconstruction based on aggregated masks uploaded by clients. The server then synchronizes the updated global model to clients for the next round of training. In Figure \ref{fig:overview_design}, we mark the 7 sequential steps in a single FL round as circled numbers to clearly show the interplay between federated training and secure aggregation protocol.

This software architecture has two special designs that can further reduce the computational and communication overhead of the secure aggregation protocol.

\begin{figure*}[h!]
\centering
    \subfigure[ Non-overlapped ]{\label{fig:pipeline_NonOverlap}
    \includegraphics[width=.47\textwidth]{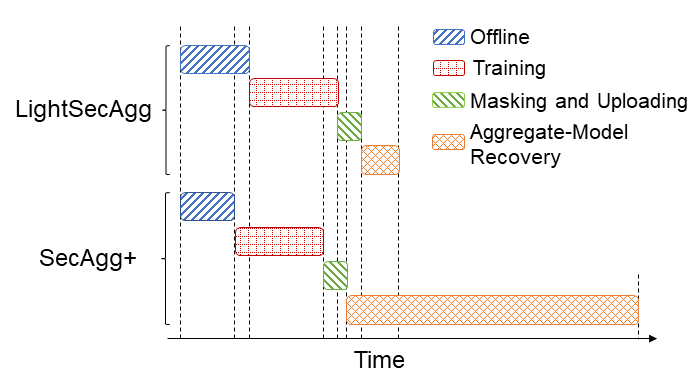}
    }
    \subfigure[ Overlapped ]{\label{fig:pipeline_Overlap}
    \includegraphics[width=.47\textwidth]{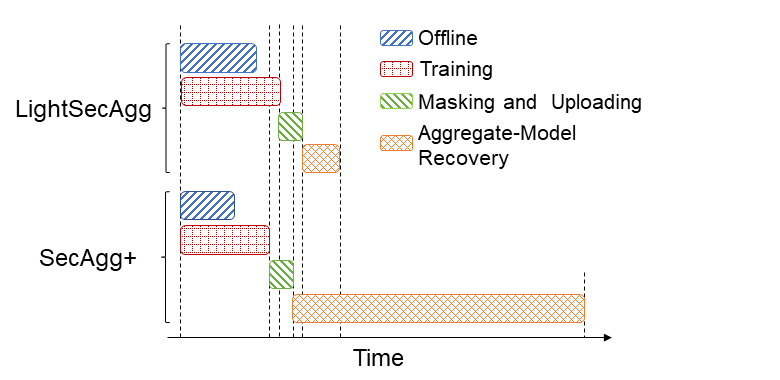}
    }
    \vspace{-4mm}
\caption{The timing diagram of the overlapped implementation in \scheme and \googlep \cite{bell2020secure} for a single FL training round to train MobileNetV3 \cite{howard2019searching} with CIFAR-100 dataset \cite{krizhevsky2009learning}. \google \cite{bonawitz2017practical} is not included as it takes much longer than other two protocols. }
\label{fig:pipeline}
\vspace{-3mm}
\end{figure*}

\vspace{-1mm}
\noindent {\bf Parallelization of offline phase and model training.} 
We note that for all considered protocols, \scheme, \google, and \googlep,
the communication and computation time to generate and exchange the random masks in the offline phase can be \emph{overlapped} with model training. Hence, in our design, we reduce the offline computation and communication overhead by allowing each user to train the model and carry out the offline phase simultaneously by running two parallel processes (multi-threading performs relatively worse due to Python GIL, Global Interpreter Lock), as shown as purple and red colors in Figure \ref{fig:overview_design}.
We also demonstrate the timing diagram of the overlapped implementation in a single FL training round in Figure \ref{fig:pipeline}. 
We will analyze its impact on overall acceleration in section \ref{label:analysis}.

\begin{table*}[t!]
\centering
\footnotesize 
\captionof{table}{Summary of four implemented machine learning tasks and performance gain of \scheme with respect to \google and \googlep. All learning tasks are for image classification. MNIST, FEMNIST and CIFAR-10 are low-resolution datasets, while images in GLD-23K are high resolution, which cost much longer training time; LR and CNN are shallow models, but MobileNetV3 and EfficientNet-B0 are much larger models, but they are tailored for efficient edge training and inference.
}
\label{tbl:summary}
\scalebox{0.85}{
\begin{tabular}[b]{|c|c|c|c|c|c|c|} 
    \hline
    \multirow{2}{*}{No.} & \multirow{2}{*}{Dataset} & \multirow{2}{*}{Model} & \multirow{2}{*}{Model Size} & \multicolumn{3}{c|}{Gain} \\
    \cline{5-7}
       &  &  & ($d$) & Non-overlapped & Overlapped & Aggregation-only \\
    \hline
    1 & MNIST~\cite{lecun1998gradient} & Logistic Regression  & $7,\!850$  & $6.7\times$, $2.5\times$ & $8.0\times$, $2.9\times$ & $13.0\times$, $4.1\times$ \\
    \hline
    2 & FEMNIST~\cite{caldas2018leaf} & CNN~\cite{mcmahan2017communication} & $1,\!206,\!590$ & $11.3\times$, $3.7\times$ & $12.7\times$, $4.1\times$ & $13.2\times$, $4.2\times$\\
    \hline
    3 & CIFAR-10~\cite{krizhevsky2009learning} & MobileNetV3~\cite{howard2019searching}       & $3,\!111,\!462$ & $7.6\times$, $2.8\times$ & $9.5\times$, $3.3\times$ & $13.1\times$, $3.9\times$\\
    \hline
    4 & GLD-23K~\cite{weyand2020google} & EfficientNet-B0~\cite{tan2019efficientnet}   & $5,\!288,\!548$ & $3.3\times$, $1.6\times$ & $3.4\times$, $1.7\times$ & $13.0\times$, $4.1\times$\\
    \hline
\end{tabular}
}\vspace{-5mm}
\end{table*}

\vspace{-1mm}
\noindent {\bf Optimized federated training system and communication APIs via tensor-aware RPC (Remote Procedure Call).} As the yellow blocks in Figure \ref{fig:overview_design} show, we specially design the sending and receiving queues to accelerate the scenario that the device has to be sender and receiver simultaneously. As such, the offline phase of \scheme can further be accelerated by parallelizing the transmission and reception of $[\mathbf{\tilde{z}}_i]_j$. 
This design can also speed up the offline pairwise agreement in \google and \googlep. Moreover, we choose PyTorch RPC \cite{trpc} as the communication backend rather than gRPC \cite{grpc} and MPI \cite{mpi} because its tensor-aware communication API can reduce the latency in scenarios where the communicator is launched frequently, i.e., each client in the offline mask exchanging phase needs to distribute $N$ coded segments to $N$ users. 

With the above design, we can deploy \scheme in both embedded IoT devices and AWS EC2 instances. AWS EC2 instances can also represent a realistic cross-device setting because, in our experiments, we use AWS EC2 \texttt{m3.medium} instances, which are CPU-based and have the same hardware configuration as modern smartphones such as iOS and Android devices. Furthermore, we package our system as a Docker image to simplify the system deployment to hundreds of edge devices.

\section{Experimental Results}\label{sec:experiment}
\begin{figure*}[t!]
    \centering
    \vspace{-3pt}
    \subfigure[Non-overlapped]{\label{fig:runtime_CNN_NonOverlap_varDropout}
    \vspace{-10pt}
    \includegraphics[width=.42\textwidth]{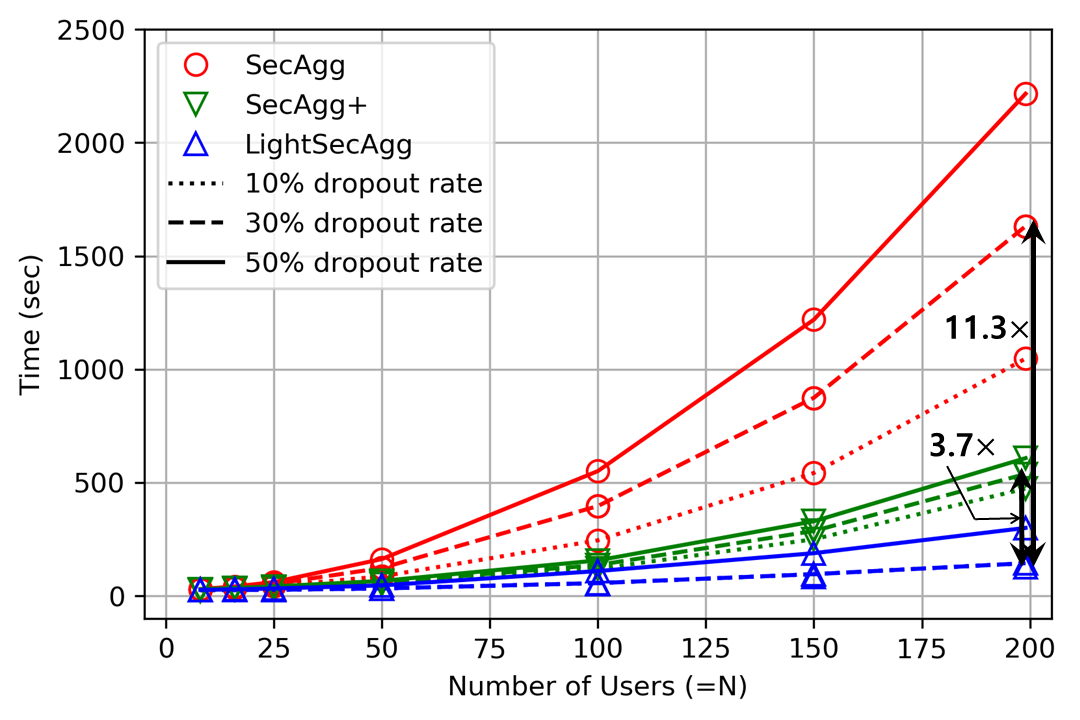}
    \vspace{-20pt}
    }
    \subfigure[Overlapped]{\label{fig:runtime_CNN_Overlap_varDropout}
    \includegraphics[width=.42\textwidth]{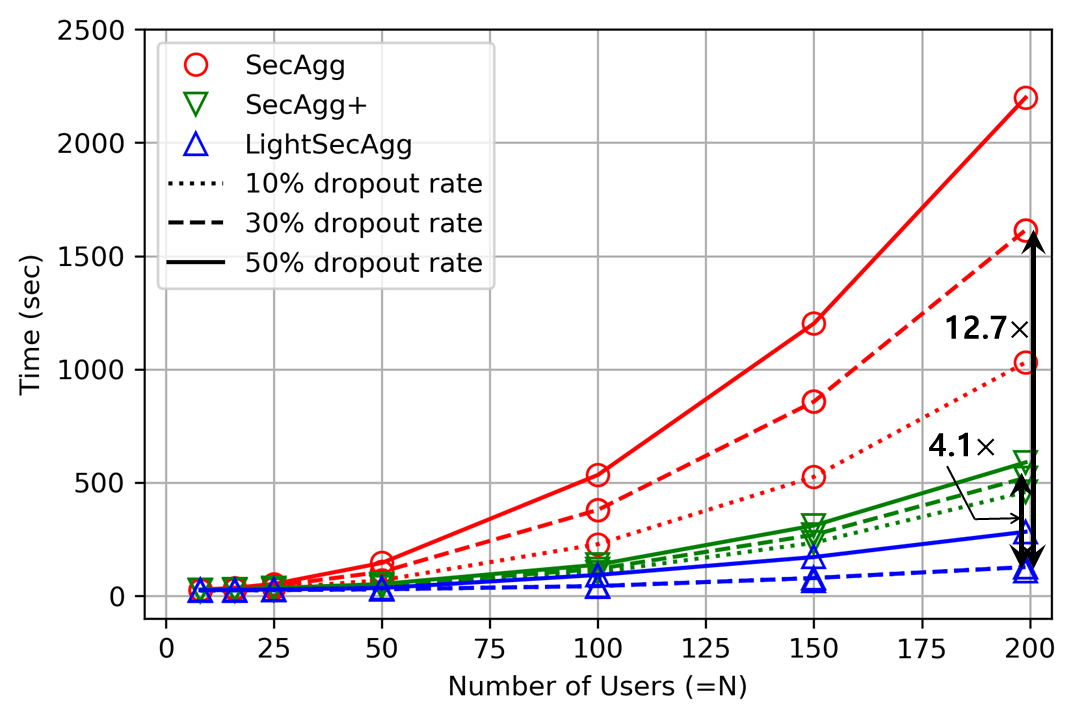}
    }
    \vspace{-4mm}
\caption{Total running time of \scheme versus the state-of-the-art protocols (\google and \googlep) to train CNN~\cite{mcmahan2017communication} on the FEMNIST dataset~\cite{caldas2018leaf}, as the number of users increases, for various dropout rates.}
\label{fig:runtime_CNN_varDropout}
\vspace{-0.2cm}
\end{figure*}

\subsection{Setup}

\noindent {\bf Dataset and models.} 
To provide a comprehensive coverage of realistic FL settings, we train four models over computer vision datasets of different sizes, summarized in Table \ref{tbl:summary}. The hyper-parameter settings are provided in Appendix \ref{appendix:exp}. 


\noindent {\bf Dropout rate.} To model the dropped users, we randomly select $pN$ users where $p$ is the dropout rate. 
We consider the worst-case scenario \cite{bonawitz2017practical}, where the selected $pN$ users artificially drop after uploading the masked model. All three protocols provide privacy guarantee $T=\frac{N}{2}$ and resiliency for three different dropout rates, $p=0.1$, $p=0.3$, and $p=0.5$, which are realistic values according to the industrial observation in real FL system \cite{bonawitz2019towards}. As we can see that when carefully selecting devices which may be stable online during the time period of training, the dropout rate is as high as 10\%; when considering intermittently connected devices, only up to 10K devices can participate simultaneously when there are 10M daily active devices ($1:1000$). 

\noindent {\bf Number of users and Communication Bandwidth.} In our experiments, we train up to $N=200$ users. The measured real bandwidth is $320$Mb/s. We also consider two other bandwidth settings of $4$G (LTE-A) and $5$G cellular networks as we discuss later.

\noindent {\bf Baselines.} We analyze and compare the performance of \scheme with two baseline schemes: \google and \googlep described in Section~\ref{sec:overviews}. While there are also other secure aggregation protocols (e.g., \turbo~\cite{so2021turbo} and \fast\cite{kadhe2020fastsecagg}), we use \google and \googlep for our baselines since other schemes weaken the privacy guarantees as we discussed in Related Works part of Section~\ref{sec:intro}.




\subsection{Overall Evaluation and Performance Analysis}
\label{label:analysis}

For the performance analysis, we measure the total running time for a single round of global iteration which includes model training and secure aggregation with each protocol while increasing the number of users $N$ gradually for different user dropouts. 
Our results from training CNN~\cite{mcmahan2017communication} on the FEMNIST dataset~\cite{caldas2018leaf} are demonstrated in Figure~\ref{fig:runtime_CNN_varDropout}. The performance gain of \scheme with respect to \google and \googlep to train the other models is also provided in Table \ref{tbl:summary}.
More detailed experimental results are provided in Appendix~\ref{appendix:exp}.
We make the following key observations.

\vspace{-1mm}
\textbf{Impact of dropout rate}: the total running time of \google and \googlep increases monotonically with the dropout rate. This is because their total running time is dominated by the mask recovery at the server, which increases quadratically with the number of users.

\vspace{-1mm}
\textbf{Non-overlapping v.s. Overlapping:} In the non-overlapped implementation, \scheme provides a speedup of up to $11.3\times$ and $3.7\times$ over \google and \googlep, respectively, by significantly reducing the server's execution time; in the overlapped implementation, \scheme provides a further speedup of up to $12.7\times$ and $4.1\times$ over \google and \googlep, respectively.
This is due to the fact that \scheme requires more communication and a higher computational cost in the offline phase than the baseline protocols, and the overlapped implementation helps to mitigate this extra cost.


\vspace{-1mm}
\textbf{Impact of model size:} \scheme provides a significant speedup of the aggregate-model recovery phase at the server over the baseline protocols in all considered model sizes. When training EfficientNet-B0 on GLD-23K dataset, \scheme provides the smallest speedup in the most training-intensive task. This is because training time is dominant in this task, and training takes almost the same time in \scheme and baseline protocols.

\textbf{Aggregation-only:}
When comparing the aggregation time only, the speedup remains the same for various model sizes as shown in Table~\ref{tbl:summary}.
We note that speeding up the aggregation phase by itself is still very important because local training and aggregation phases are not necessarily happening one immediately after the other.
For example, local training may be done sporadically and opportunistically throughout the day (whenever resources are available), while global aggregation may be postponed to a later time when a large fraction of the users are done with local training, and they are available for aggregation (e.g., $2$ am).
    
\vspace{-1mm}
\textbf{Impact of $U$:} \scheme incurs the smallest running time for the case when $p=0.3$, which is almost identical to the case when $p=0.1$. Recall that \scheme can select the design parameter $U$ between $T = 0.5N$ and $N-D = (1-p)N$. Within this range, while increasing $U$ reduces the size of the symbol to be decoded, it also increases the complexity of decoding each symbol. The experimental results suggest that the optimal choices for the cases of $p=0.1$ and $p=0.3$ are both $U=\lfloor0.7N\rfloor$, which leads to a faster execution than when $p=0.5$, where $U$ can only be chosen as $U=0.5N+1$.

\begin{table}[h!]
\centering
\vspace{-0.5cm}
\caption{Performance gain in different bandwidth settings.}
    \footnotesize
    \begin{tabular}{|c|c|c|c|}
    \hline 
        Protocols  & $4$G ($98$ Mbps) & $320$ Mbps  & $5$G ($802$ Mbps) \\
    \hline
    \google   & $8.5\times$ & $12.7\times$ & $13.5\times$ \\
    \hline
    \googlep  & $2.9\times$ & $4.1\times$  & $4.4\times$ \\
    \hline
    \end{tabular}
    \label{tbl:gain_variousNW}
\vspace{-0.3cm}
\end{table}

\textbf{Impact of Bandwidth:}
We have also analyzed the impact of communication bandwidth at the users. In addition to the default bandwidth setting used in this section, we have considered two other edge scenarios: $4$G (LTE-A) and $5$G cellular networks using realistic bandwidth settings of $98$ and $802$ Mbps respectively~\cite{minovski2021throughput, scheuner2018cloud}). The results are reported in Table \ref{tbl:gain_variousNW} for a single FL round to train CNN over FEMNIST.

\begin{table*}[t!]
\centering
\footnotesize 
\captionof{table}{Breakdown of the running time (sec) of \scheme and the state-of-the-art protocols (\google \cite{bonawitz2017practical} and \googlep~\cite{bell2020secure}) to train CNN~\cite{mcmahan2017communication} on the FEMNIST dataset~\cite{caldas2018leaf} with $N=200$ users, for dropout rate $p = 10\%, 30\%, 50\%$.
}
\label{tbl:breakdown_EfficientNet}
\scalebox{0.9}{
\begin{tabular}[b]{|c|c|c|c|c|c|c|c|} 
    \hline
    \multirow{2}{*}{Protocols} & \multirow{2}{*}{Phase} & \multicolumn{3}{c|}{Non-overlapped} & \multicolumn{3}{c|}{Overlapped}\\
    \cline{3-8}
       &  & $p=10\%$ & $p=30\%$ & $p=50\%$ & $p=10\%$ & $p=30\%$ & $p=50\%$ \\
   \hline
   \multirow{5}{*}{\scheme} & Offline & $69.3$ & $69.0$ & $191.2$ & \multirow{2}{*}{$75.1$} & \multirow{2}{*}{$74.9$} & \multirow{2}{*}{$196.9$}\\ \cline{2-5}
    & Training  & $22.8$  & $22.8$ & $22.8$ & & & \\ \cline{2-8}
    & Uploading & $12.4$  & $12.2$ & $21.6$ & $12.6$  & $12.0$ & $21.4$ \\ \cline{2-8}
    & Recovery  & $40.9$  & $40.7$ & $64.5$ & $40.7$  & $41.0$ & $64.9$ \\ \cline{2-8}
    & Total     & $145.4$ & $144.7$ & $300.1$ & $123.4$ & $127.3$ & $283.2$ \\ \hline
    
    \multirow{5}{*}{\google} & Offline & $95.6$ & $98.6$ & $102.6$& \multirow{2}{*}{$101.2$}& \multirow{2}{*}{$102.3$}& \multirow{2}{*}{$101.3$}\\ \cline{2-5}
    & Training  & $22.8$  & $22.8$ & $22.8$ & & & \\ \cline{2-8}
    & Uploading & $10.7$ & $10.9$ & $11.0$ & $10.9$ & $10.8$ & $11.2$\\ \cline{2-8}
    & Recovery  & $911.4$ & $1499.2$ & $2087.0$ & $911.2$ & $1501.3$ & $2086.8$\\ \cline{2-8}
    & Total     & $1047.5$& $1631.5$ & $2216.4$ & $1030.3$ & $1614.4$ & $2198.9$\\ \hline
    
    \multirow{5}{*}{\googlep} & Offline & $67.9$ & $68.1$& $69.2$& \multirow{2}{*}{$73.9$} & \multirow{2}{*}{$73.8$} & \multirow{2}{*}{$74.2$}\\ \cline{2-5}
    & Training  & $22.8$  & $22.8$ & $22.8$ & & & \\ \cline{2-8}
    & Uploading & $10.7$ & $10.8$ & $10.7$ & $10.7$ & $10.8$ & $10.9$\\ \cline{2-8}
    & Recovery  & $379.1$ & $436.7$ & $495.5$ & $378.9$ & $436.7$ & $497.3$\\ \cline{2-8}
    & Total     & $470.5$ & $538.4$ & $608.2$ & $463.6$ & $521.3$ & $582.4$\\ \hline

\end{tabular}
\label{table:runtime_breakdown}
\vspace{-0.3cm}
}
\end{table*}

\subsection{Performance Breakdown}

To further investigate the primary gain of \scheme, we provide the breakdown of total running time for training CNN~\cite{mcmahan2017communication} on the FEMNIST dataset~\cite{caldas2018leaf} in Table~\ref{table:runtime_breakdown}. The breakdown of the running time confirms that the primary gain lies in the complexity reduction at the server provided by \scheme, especially for a large number of users.

\vspace{-0.2cm}
\subsection{Convergence Performance in Asynchronous FL} 
\vspace{-0.3cm}

\begin{figure}[h!]
\centering
    {
    \includegraphics[width=.38\textwidth]{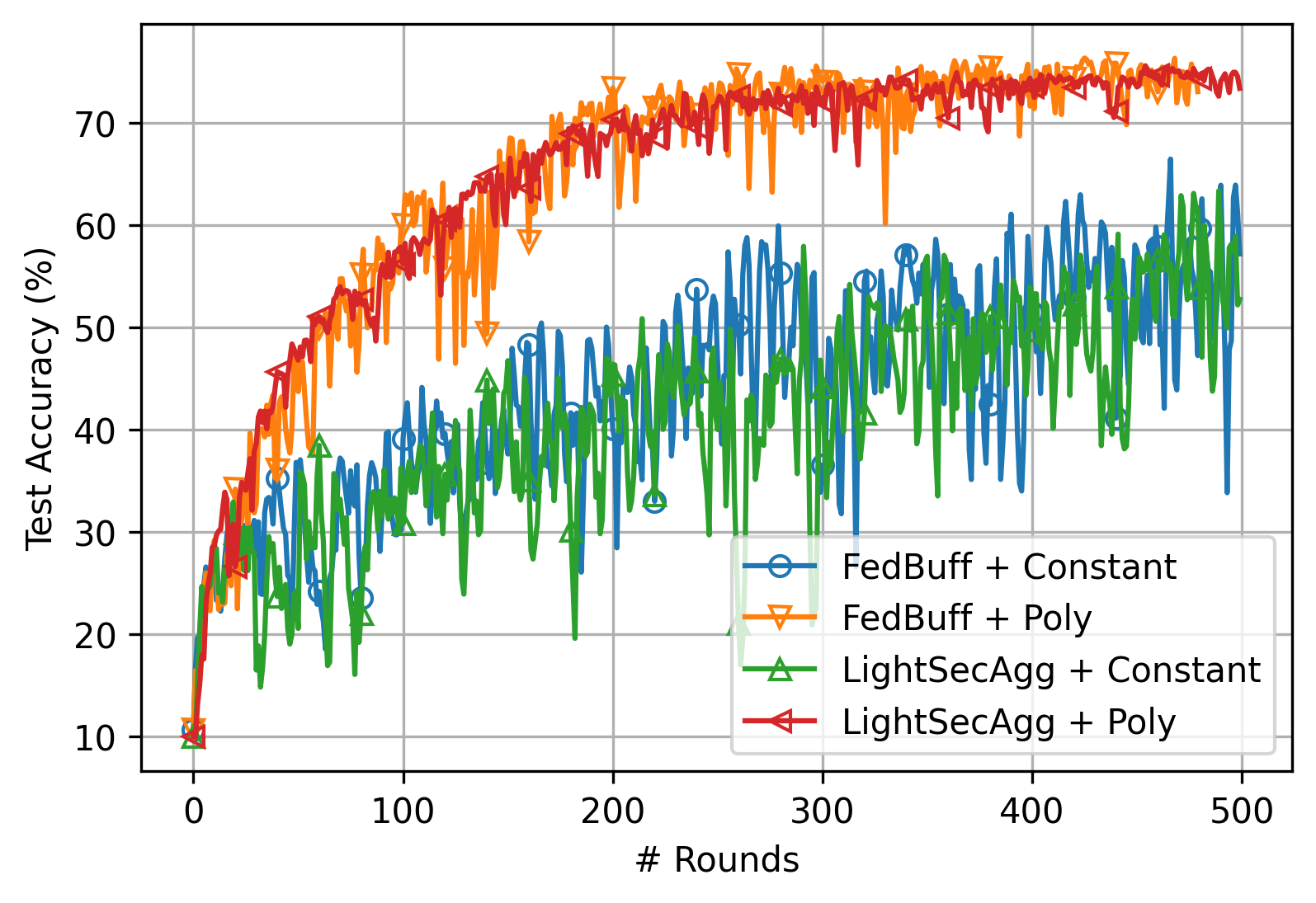}
    }
    \vspace{-12pt}
\caption{\footnotesize Accuracy of asynchronous \scheme and \FedBuff on CIFAR-10 dataset~\cite{krizhevsky2009learning} with two strategies for mitigating the staleness: a constant function $s(\tau)=1$ named \emph{Constant}; and a polynomial function $s_\alpha(\tau)=(1+\tau)^{-\alpha}$ named \emph{Poly} where $\alpha=1$. The accuracy is reasonable since we use a variant of LeNet-5~\cite{xie2019asynchronous}.}
\label{fig:async_convergence}
\vspace{-0.2cm}
\end{figure}

As described in Remark~\ref{rmk:Incompatibility}, \google and \googlep are not applicable to asynchronous FL, and hence we cannot compare the total running time of \scheme with these baseline secure aggregation protocols.
As such, in our experiments here we instead focus on convergence performance of \scheme compared to \FedBuff \cite{nguyen2021federated} to investigate the impact of asynchrony and quantization in performance. In Figure \ref{fig:async_convergence}, we demonstrate that \scheme has almost the same performance as \FedBuff on CIFAR-10 dataset while \scheme includes quantization noise to protect the privacy of individual local updates of users.
The details of the experiment setting and additional experiments for asynchronous FL are provided in Appendix~\ref{app:sub:async-exp}.

\vspace{-0.3cm}
\section{Conclusion and Future Works}\label{sec:conclusion}
\vspace{-0.2cm}
This paper proposed \scheme, a new approach for secure aggregation in synchronous and asynchronous FL. Compared with the state-of-the-art protocols, \scheme reduces the overhead of model aggregation in FL by leveraging one-shot aggregate-mask reconstruction of the surviving users, while providing the same privacy and dropout-resiliency guarantees. In a realistic FL framework, via extensive empirical results it is also shown that \scheme can provide substantial speedup over baseline protocols for training diverse machine learning models. While we focused on privacy in this work (under the honest but curious threat model), an interesting future research is to combine \scheme with state-of-the-art Byzantine robust aggregation protocols (e.g.,~\cite{he2020secure, BREA,elkordy2021basil,karimireddy2021learning}) to also mitigate Byzantine users while ensuring privacy. 


\clearpage

\bibliographystyle{mlsys2022}
\bibliography{main}

\clearpage
\onecolumn
\appendix
\section*{Appendix}
\section{Pseudo Code of \scheme}\label{appendix:code}
\begin{algorithm}
	\caption{The \scheme protocol} \label{alg:lightsecagg}
\textbf{Input:}  $T$ (privacy guarantee), $D$ (dropout-resiliency guarantee), $U$ (target number of surviving users)
\begin{algorithmic}[1]
\STATE \textbf{Server Executes:}
\STATE {\color{red}// \emph{phase: offline encoding and sharing of local masks}}
\FOR {each user $i= 1,2,\ldots,N$ \textbf{in parallel}}
    \STATE $\mathbf{z}_i\gets$ randomly picks from $\mathbb{F}^d_q$
    \STATE $[\mathbf{z}_i]_1,\ldots,[\mathbf{z}_i]_{U-T}\gets$ obtained by partitioning $\mathbf{z}_i$ to $U-T$ pieces
    \STATE $[\mathbf{n}_i]_{U-T+1},\ldots,[\mathbf{n}_i]_{U}\gets$ randomly picks from $\mathbb{F}^{\frac{d}{U-T}}_q$  
    \STATE $\{[\mathbf{\tilde{z}}_i]_j\}_{j \in [N]}\gets$ obtained by encoding $[\mathbf{z}_i]_k$'s and $[\mathbf{n}_i]_k$'s using \eqref{eq:def_z_tilde}
   \STATE sends encoded mask $[\mathbf{\tilde{z}}_i]_j$ to user $j \in [N]\backslash \{i\}$
   \STATE receives encoded mask $[\mathbf{\tilde{z}}_j]_i$ from user $j \in [N]\backslash \{i\}$
\ENDFOR
\STATE {\color{red}// \emph{phase: masking and uploading of local models}}
\FOR {each user $i= 1,2,\ldots,N$ \textbf{in parallel}}
\STATE // \emph{user $i$ obtains $\mathbf{x}_i$ after the local update}\\
\STATE $\mathbf{\tilde{x}}_i \gets \mathbf{x}_i + \mathbf{z}_i$ // masks the local model
\STATE uploads masked model $\mathbf{\tilde{x}}_i$ to the server
\ENDFOR
\STATE identifies set of surviving users $\mathcal{U}_1 \subseteq [N]$
\STATE gathers masked models $\mathbf{\tilde{x}}_i$ from user $i \in \mathcal{U}_1$
\STATE {\color{red}// \emph{phase: one-shot aggregate-model recovery}}
\FOR {each user $i \in \mathcal{U}_1$ \textbf{in parallel}}
\STATE computes aggregated encoded masks $ \sum_{j \in \mathcal{U}_1}[\mathbf{\tilde{z}}_j]_i$
\STATE uploads aggregated encoded masks $ \sum_{j \in \mathcal{U}_1}[\mathbf{\tilde{z}}_j]_i$ to the server
\ENDFOR
\STATE collects $U$ messages of aggregated encoded masks $ \sum_{j \in \mathcal{U}_1}[\mathbf{\tilde{z}}_j]_i$ from user $i \in \mathcal{U}_1$
\STATE // \emph{recovers the aggregated-mask} \\
\STATE $\sum_{i\in \mathcal{U}_1}\mathbf{z}_i\gets$ obtained by decoding the received $U$ messages 
\STATE // \emph{recovers the aggregate-model for the surviving users}\\ 
\STATE $\sum_{i\in \mathcal{U}_1}\mathbf{x}_i\gets\sum_{i\in \mathcal{U}_1}\mathbf{\tilde{x}}_i - \sum_{i\in \mathcal{U}_1}\mathbf{z}_i$ 
\end{algorithmic}
\end{algorithm}

\section{Proof of Theorem \ref{thm:main_thm}}\label{appendix:main_thm}
We prove the dropout-resiliency guarantee and the privacy guarantee for a single FL training round. As all randomness is independently generated across each round, one can extend the dropout-resiliency guarantee and the privacy guarantee for all training rounds for both synchronous and asynchronous FL setting. For simplicity, round index $t$ is omitted in this proof.

For any pair of privacy guarantee $T$ and dropout-resiliency guarantee $D$ such that $T+D<N$, we select an arbitrary $U$ such that $N-D\geq U > T$. In the following, we show that \scheme with chosen design parameters $T$, $D$ and $U$ can simultaneously achieve (1) privacy guarantee against up to any $T$ colluding users, and (2) dropout-resiliency guarantee against up to any $D$ dropped users. We denote the concatenation of $\{[\mathbf{n}_i]_k\}_{k\in {U-T+1,\dots,U}}$ by $\mathbf{n}_i$ for $i \in [N]$.

{\bf (Dropout-resiliency guarantee)} We now focus on the phase of one-shot aggregate-model recovery. Since each user encodes its sub-masks by the same MDS matrix $W$, each $\sum_{i\in \mathcal{U}_1}[\mathbf{\tilde{z}}_i]_j$ is an encoded version of $\sum_{i \in \mathcal{U}_1}[\mathbf{z}_i]_k$ for $k \in [U-T]$ and $\sum_{i \in \mathcal{U}_1}[\mathbf{n}_i]_k$ for $k \in \{U-T+1,\dots,U\}$ as follows:
\begin{align}
    \sum_{i \in \mathcal{U}_1}[\mathbf{\tilde{z}}_i]_j =( \sum_{i \in \mathcal{U}_1}[\mathbf{z}_i]_1,\dots,  \sum_{i \in \mathcal{U}_1}[\mathbf{z}_i]_{U-T}, \sum_{i \in \mathcal{U}_1}[\mathbf{n}_i]_{U-T+1},\dots , \sum_{i \in \mathcal{U}_1}[\mathbf{n}_i]_U)\cdot W_j,
\end{align}
where $W_j$ is the $j$'th column of $W$.

Since $N-D\geq U$, there are at least $U$ surviving users after user dropouts. Thus, the server is able to recover $\sum_{i \in \mathcal{U}_1}[\mathbf{z}_i]_k$ for $k \in [U-T]$ via MDS decoding after receiving a set of any $U$ messages from the surviving users. Recall that $[\mathbf{z}_i]_k$'s are sub-masks of $\mathbf{z}_i$, so the server can successfully recover $\sum_{i\in \mathcal{U}_1}\mathbf{z}_i$. Lastly, the server recovers the aggregate-model for the set of surviving users $\mathcal{U}_1$ by $\sum_{i \in \mathcal{U}_1}\mathbf{x}_i = \sum_{i \in \mathcal{U}_1}\mathbf{\tilde{x}}_i - \sum_{i \in \mathcal{U}_1}\mathbf{z}_i= \sum_{i \in \mathcal{U}_1}(\mathbf{x}_i + \mathbf{z}_i) - \sum_{i \in \mathcal{U}_1}\mathbf{z}_i$.


{\bf (Privacy guarantee)} We first present Lemma~\ref{lemma}, whose proof is provided in Appendix~\ref{appendix:lemma}.
\begin{lemma}\label{lemma}\normalfont
For any $\mathcal{T} \subseteq [N]$ of size $T$ and any $\mathcal{U}_1 \subseteq [N]$, $|\mathcal{U}_1|\geq U$ such that $U>T$, if the random masks $[\mathbf{n}_i]_k$'s are jointly uniformly random, we have
\begin{align}
    & I(\{\mathbf{z}_i\}_{i\in[N]\backslash \mathcal{T}};\{\mathbf{z}_i\}_{i\in \mathcal{T}},\{[\mathbf{\tilde{z}}_j]_i\}_{j\in [N], i \in \mathcal{T}}) = 0.
\end{align}
\end{lemma}

We consider the worst-case scenario in which all the messages sent from the users are received by the server during the execution of \scheme, i.e., the users identified as dropped are delayed. Thus, the server receives $\mathbf{x}_i+\mathbf{z}_i$ from user $i \in [N]$ and $\sum_{j\in \mathcal{U}_1}[\mathbf{\tilde{z}}_j]_i$ from user $i\in \mathcal{U}_1$. 
We now show that \scheme provides privacy guarantee $T$, i.e., for an arbitrary set of colluding users $\mathcal{T}$ of size $T$, the following holds,
\begin{align}
    I\left(\{\mathbf{x}_i\}_{i \in [N]};\{\mathbf{x}_i+\mathbf{z}_i\}_{i \in [N]}, \{\sum_{j\in \mathcal{U}_1}[\mathbf{\tilde{z}}_j]_i\}_{i \in \mathcal{U}_1}\Bigg|\sum_{i \in \mathcal{U}_1}\mathbf{x}_i,\{\mathbf{x}_i\}_{i \in \mathcal{T}},\{\mathbf{z}_i\}_{i \in \mathcal{T}}, \{[\mathbf{\tilde{z}}_j]_i\}_{j\in [N], i \in \mathcal{T}}\right) =0.
\end{align}

We prove it as follows:
\begin{align}
  & I\left(\{\mathbf{x}_i\}_{i \in [N]};\{\mathbf{x}_i+\mathbf{z}_i\}_{i \in [N]}, \{\sum_{j\in \mathcal{U}_1}[\mathbf{\tilde{z}}_j]_i\}_{i \in \mathcal{U}_1}\Bigg|\sum_{i \in \mathcal{U}_1}\mathbf{x}_i,\{\mathbf{x}_i\}_{i \in \mathcal{T}},\{\mathbf{z}_i\}_{i \in \mathcal{T}}, \{[\mathbf{\tilde{z}}_j]_i\}_{j\in [N], i \in \mathcal{T}}\right)  \\ 
  = & H\left(\{\mathbf{x}_i+\mathbf{z}_i\}_{i \in [N]}, \{\sum_{j\in \mathcal{U}_1}[\mathbf{\tilde{z}}_j]_i\}_{i \in \mathcal{U}_1}\Bigg|\sum_{i \in \mathcal{U}_1}\mathbf{x}_i,\{\mathbf{x}_i\}_{i \in \mathcal{T}},\{\mathbf{z}_i\}_{i \in \mathcal{T}}, \{[\mathbf{\tilde{z}}_j]_i\}_{j\in [N], i \in \mathcal{T}} \right) \nonumber\\
  & - H\left(\{\mathbf{x}_i+\mathbf{z}_i\}_{i \in [N]}, \{\sum_{j\in \mathcal{U}_1}[\mathbf{\tilde{z}}_j]_i\}_{i \in \mathcal{U}_1}\Bigg|\{\mathbf{x}_i\}_{i \in [N]},\{\mathbf{z}_i\}_{i \in \mathcal{T}}, \{[\mathbf{\tilde{z}}_j]_i\}_{j\in [N], i \in \mathcal{T}} \right) \\
  = & H\left(\{\mathbf{x}_i+\mathbf{z}_i\}_{i \in [N]}, \sum_{i\in \mathcal{U}_1}\mathbf{z}_i ,\sum_{i\in \mathcal{U}_1}\mathbf{n}_i \Bigg|\sum_{i \in \mathcal{U}_1}\mathbf{x}_i,\{\mathbf{x}_i\}_{i \in \mathcal{T}},\{\mathbf{z}_i\}_{i \in \mathcal{T}}, \{[\mathbf{\tilde{z}}_j]_i\}_{j\in [N], i \in \mathcal{T}} \right) \nonumber\\
  & - H\left(\{\mathbf{z}_i\}_{i \in [N]},\sum_{i\in \mathcal{U}_1}\mathbf{z}_i ,\sum_{i\in \mathcal{U}_1}\mathbf{n}_i\Bigg|\{\mathbf{x}_i\}_{i \in [N]},\{\mathbf{z}_i\}_{i \in \mathcal{T}}, \{[\mathbf{\tilde{z}}_j]_i\}_{j\in [N], i \in \mathcal{T}} \right) \label{eq:privacy6}\\
  = & H\left(\{\mathbf{x}_i+\mathbf{z}_i\}_{i \in [N] \backslash \mathcal{T}}, \sum_{i\in \mathcal{U}_1}\mathbf{z}_i ,\sum_{i\in \mathcal{U}_1}\mathbf{n}_i \Bigg|\sum_{i \in \mathcal{U}_1}\mathbf{x}_i,\{\mathbf{x}_i\}_{i \in \mathcal{T}},\{\mathbf{z}_i\}_{i \in \mathcal{T}}, \{[\mathbf{\tilde{z}}_j]_i\}_{j\in [N], i \in \mathcal{T}} \right) \nonumber\\
  & - H\left(\{\mathbf{z}_i\}_{i \in [N]},\sum_{i\in \mathcal{U}_1}\mathbf{z}_i ,\sum_{i\in \mathcal{U}_1}\mathbf{n}_i\Bigg|\{\mathbf{x}_i\}_{i \in [N]},\{\mathbf{z}_i\}_{i \in \mathcal{T}}, \{[\mathbf{\tilde{z}}_j]_i\}_{j\in [N], i \in \mathcal{T}} \right) \label{eq:privacy7}\\
  = & H\left(\{\mathbf{x}_i+\mathbf{z}_i\}_{i \in [N] \backslash \mathcal{T}}  \Bigg|\sum_{i \in \mathcal{U}_1}\mathbf{x}_i,\{\mathbf{x}_i\}_{i \in \mathcal{T}},\{\mathbf{z}_i\}_{i \in \mathcal{T}}, \{[\mathbf{\tilde{z}}_j]_i\}_{j\in [N], i \in \mathcal{T}} \right) \nonumber\\
    & + H\left( \sum_{i\in \mathcal{U}_1}\mathbf{z}_i ,\sum_{i\in \mathcal{U}_1}\mathbf{n}_i \Bigg|\{\mathbf{x}_i+\mathbf{z}_i\}_{i \in [N] \backslash \mathcal{T}},\sum_{i \in \mathcal{U}_1}\mathbf{x}_i,\{\mathbf{x}_i\}_{i \in \mathcal{T}},\{\mathbf{z}_i\}_{i \in \mathcal{T}}, \{[\mathbf{\tilde{z}}_j]_i\}_{j\in [N], i \in \mathcal{T}} \right) \nonumber\\
    & - H\left(\{\mathbf{z}_i\}_{i \in [N]}\Bigg|\{\mathbf{x}_i\}_{i \in [N]},\{\mathbf{z}_i\}_{i \in \mathcal{T}}, \{[\mathbf{\tilde{z}}_j]_i\}_{j\in [N], i \in \mathcal{T}} \right) \nonumber \\
    & - H\left( \sum_{i\in \mathcal{U}_1}\mathbf{z}_i ,\sum_{i\in \mathcal{U}_1}\mathbf{n}_i\Bigg|\{\mathbf{z}_i\}_{i \in [N]},\{\mathbf{x}_i\}_{i \in [N]},\{\mathbf{z}_i\}_{i \in \mathcal{T}}, \{[\mathbf{\tilde{z}}_j]_i\}_{j\in [N], i \in \mathcal{T}} \right) \label{eq:privacy8}\\
      = & H\left(\{\mathbf{x}_i+\mathbf{z}_i\}_{i \in [N] \backslash \mathcal{T}}  \Bigg|\sum_{i \in \mathcal{U}_1}\mathbf{x}_i,\{\mathbf{x}_i\}_{i \in \mathcal{T}},\{\mathbf{z}_i\}_{i \in \mathcal{T}}, \{[\mathbf{\tilde{z}}_j]_i\}_{j\in [N], i \in \mathcal{T}} \right) \nonumber\\
    & + H\left( \sum_{i\in \mathcal{U}_1}\mathbf{n}_i \Bigg|\{\mathbf{x}_i+\mathbf{z}_i\}_{i \in [N] \backslash \mathcal{T}},\sum_{i \in \mathcal{U}_1}\mathbf{x}_i,\{\mathbf{x}_i\}_{i \in \mathcal{T}},\{\mathbf{z}_i\}_{i \in \mathcal{T}}, \{[\mathbf{\tilde{z}}_j]_i\}_{j\in [N], i \in \mathcal{T}} \right) \nonumber\\
    & - H\left(\{\mathbf{z}_i\}_{i \in [N]\backslash \mathcal{T}}\Bigg|\{\mathbf{z}_i\}_{i \in \mathcal{T}}, \{[\mathbf{\tilde{z}}_j]_i\}_{j\in [N], i \in \mathcal{T}} \right) - H\left( \sum_{i\in \mathcal{U}_1}\mathbf{n}_i\Bigg|\{\mathbf{z}_i\}_{i \in [N]}, \{[\mathbf{\tilde{z}}_j]_i\}_{j\in [N], i \in \mathcal{T}} \right) \label{eq:privacy9}\\
          = & H\left(\{\mathbf{x}_i+\mathbf{z}_i\}_{i \in [N] \backslash \mathcal{T}}  \Bigg|\sum_{i \in \mathcal{U}_1}\mathbf{x}_i,\{\mathbf{x}_i\}_{i \in \mathcal{T}},\{\mathbf{z}_i\}_{i \in \mathcal{T}}, \{[\mathbf{\tilde{z}}_j]_i\}_{j\in [N], i \in \mathcal{T}} \right) \nonumber\\
    & + H\left( \sum_{i\in \mathcal{U}_1}\mathbf{n}_i \Bigg|\{\mathbf{x}_i+\mathbf{z}_i\}_{i \in [N] \backslash \mathcal{T}},\sum_{i \in \mathcal{U}_1}\mathbf{x}_i,\{\mathbf{x}_i\}_{i \in \mathcal{T}},\{\mathbf{z}_i\}_{i \in \mathcal{T}}, \{[\mathbf{\tilde{z}}_j]_i\}_{j\in [N], i \in \mathcal{T}} \right) \nonumber\\
    & - H\left(\{\mathbf{z}_i\}_{i \in [N]\backslash \mathcal{T}}\right) - H\left( \sum_{i\in \mathcal{U}_1}\mathbf{n}_i\Bigg|\{\mathbf{z}_i\}_{i \in [N]}, \{[\mathbf{\tilde{z}}_j]_i\}_{j\in [N], i \in \mathcal{T}} \right) \label{eq:privacy16}\\
    = & 0, \label{eq:privacy10}
\end{align}
where \eqref{eq:privacy6} follows from the fact that $\{\sum_{j \in \mathcal{U}_1}[\mathbf{\tilde{z}}_j]_i\}_{i\in\mathcal{U}_1}$ is invertible to $\sum_{i\in \mathcal{U}_1}\mathbf{z}_i$ and $\sum_{i\in \mathcal{U}_1}\mathbf{n}_i$. Equation \eqref{eq:privacy7} holds since $\{\mathbf{x}_i + \mathbf{z}_i\}_{i \in \mathcal{T}}$ is a deterministic function of  $\{\mathbf{z}_i\}_{i \in \mathcal{T}}$ and $\{\mathbf{x}_i\}_{i \in \mathcal{T}}$. Equation \eqref{eq:privacy8} follows from the chain rule. In equation \eqref{eq:privacy9}, the second term follows from the fact that $\sum_{i\in \mathcal{U}_1}\mathbf{z}_i$ is a deterministic function of $\{\mathbf{x}_i+\mathbf{z}_i\}_{i \in [N] \backslash \mathcal{T}}$, $\sum_{i \in \mathcal{U}_1}\mathbf{x}_i$, $\{\mathbf{x}_i\}_{i \in \mathcal{T}}$,$\{\mathbf{z}_i\}_{i \in \mathcal{T}}$; the third term follows from the independence of $\mathbf{x}_i$'s and $\mathbf{z}_i$'s; the last term follows from the fact that $\sum_{i \in \mathcal{U}_1}\mathbf{z}_i$ is a deterministic function of $\{\mathbf{z}_i\}_{i\in [N]}$ and the independence of $\mathbf{n}_i$'s and $\mathbf{x}_i$'s. In equation \eqref{eq:privacy16}, the third term follows from Lemma~\ref{lemma}. Equation~\eqref{eq:privacy10} follows from 1) $\sum_{i \in \mathcal{U}_1}\mathbf{n}_i$ is a function of $\{\mathbf{x}_i+\mathbf{z}_i\}_{i \in [N] \backslash \mathcal{T}}$, $\sum_{i \in \mathcal{U}_1}\mathbf{x}_i$ ,$\{\mathbf{x}_i\}_{i \in \mathcal{T}}$, $\{\mathbf{z}_i\}_{i \in \mathcal{T}}$ and $\{[\mathbf{\tilde{z}}_j]_i\}_{j\in [N], i \in \mathcal{T}}$; 2) $\sum_{i \in \mathcal{U}_1}\mathbf{n}_i$ is a function of $\{\mathbf{z}_i\}_{i \in \mathcal{U}_1}$ $\{[\mathbf{\tilde{z}}_j]_i\}_{j\in \mathcal{U}_1, i \in \mathcal{T}}$; 3) $\mathbf{z}_{i}$ is uniformly distributed and hence it has the maximum entropy in $\mathbb{F}^d_q$, combined with the non-negativity of mutual information.

\section{Discussion}\label{appendix:discussion}
As shown in Table~\ref{table:compare_appendix}, compared with the \google protocol~\cite{bonawitz2017practical}, \scheme significantly improves the computational efficiency at the server during aggregation. While \google requires the server to retrieve $T+1$ secret shares of a secret key for \emph{each} of the $N$ users, and to compute a single PRG function if the user survives, or $N-1$ PRG functions to recover $N-1$ pairwise masks if the user drops off, yielding a total computational load of $O(N^2d)$ at the server. In contrast, as we have analyzed in Section~\ref{sec:complexity}, for $U=O(N)$, \scheme incurs an almost constant ($O(d\log{N}$)) computational load at the server. This admits a scalable design and is expected to achieve a much faster end-to-end execution for a large number of users, given the fact that the overall execution time is dominated by the server's computation in \google~\cite{bonawitz2017practical,MLSYS2019_bd686fd6}. \google has a smaller storage overhead than \scheme as secret shares of keys with small sizes (e.g., as small as an integer) are stored, and the model size $d$ is much larger than the number of users $N$ in typical FL scenarios. This effect will also allow \google to have a smaller communication load in the phase of aggregate-model recovery. Finally, we would like to note that another advantage of \scheme over \google is the reduced dependence on cryptographic primitives such as public key infrastructure and key agreement mechanism, which further simplifies the implementation of the protocol. \googlep~\cite{bell2020secure} improves both communication and computational load of \google by considering a sparse random graph of degree $O(\log{N})$, and the complexity is reduced by factor of $O(\frac{N}{\log{N}})$. However, \googlep still incurs $O(dN\log{N})$ computational load at the server, which is much larger than $O(d\log{N})$ computational load at the server in \scheme when $U=O(N)$.
\renewcommand{\arraystretch}{1.5}
\begin{table}[htbp]
\centering
\caption{Complexity comparison between \google~\cite{bonawitz2017practical}, \googlep~\cite{bell2020secure}, and \scheme. Here $N$ is the total number of users. The parameters $d$ and $s$ respectively represent the model size and the length of the secret keys as the seeds for PRG, where $s \ll d$. \scheme and \google provide \emph{worst-case} privacy guarantee $T$ and dropout-resiliency guarantee $D$ for any $T$ and $D$ as long as $T+D<N$. \googlep provides \emph{probabilistic} privacy guarantee $T$ and dropout-resiliency guarantee $D$. \scheme selects three design parameters $T$, $D$ and $U$ such that $T<U\leq N-D$. } 
 \vspace{1mm}
\footnotesize

  \begin{tabular}{|c|c|c|c|}
 \hline 
& \google & \googlep & \scheme  \\
    \hline
 Offline storage per user &  $O(d + Ns)$ & $O(d+s\log{N})$ &  $O(d+\frac{N}{U-T}d)$     \\
     \hline
Offline communication per user & $O(sN)$ & $O(s\log{N})$ & $O(d\frac{N}{U-T})$  \\ 
    \hline
Offline computation per user & $O(dN + sN^2)$ & $O(d\log{N}+s\log^2 N)$ & $O(d\frac{N\log{N}}{U-T})$  \\ 
    \hline
Online communication per user & $O(d + sN)$& $O(d+s\log{N})$ &  $O(d + \frac{d}{U-T})$ \\ 
    \hline
Online communication at server & $O(dN + sN^2)$ & $O(dN+sN\log{N})$&  $O(dN + d\frac{U}{U-T})$ \\ 
    \hline
Online computation per user & $O(d)$ & $O(d)$ & $O(d+d\frac{U}{U-T})$ \\ 
    \hline
Decoding complexity at server & $O(s N^2)$ & $O(sN\log^2{N})$ & $O(d\frac{U \log U}{U-T} )$  \\
    \hline
PRG complexity at server &$O(dN^2)$& $O(dN\log{N})$& $-$  \\
    \hline

  \end{tabular}
  \label{table:compare_appendix}
\end{table}
\begin{table}[htbp]
\centering
\caption{Comparison of storage cost (in the number of symbols in $\mathbb{F}_q^{\frac{d}{U-T}}$) between protocol in~\cite{zhao2021information} and \scheme.}
\vspace{1mm}
  \begin{tabular}{|c|c|c|}
  \hline 
& Protocol in \cite{zhao2021information} & \scheme  \\
           \hline
Total amount of randomness needed & $N(U-T) + T\sum_{u = U}^N {N \choose u}$ &  $NU$  \\
    \hline 
Offline storage per user & $U-T + \sum_{u = U}^N {N \choose u} \frac{u}{N}$ &  $U-T+N$\\ 
    \hline
  \end{tabular}
  \label{table:compare_hua}
\end{table}

Compared with a recently proposed secure aggregation protocol in~\cite{zhao2021information}, \scheme achieves similar complexities in communication and computation during the aggregation process. The main advantage of \scheme over the scheme in~\cite{zhao2021information} lies in how the randomness is generated and stored offline at the users and the resulting reduced storage cost. For the scheme in~\cite{zhao2021information}, all randomness are generated at some external trusted party, and for each subset of ${\cal U}_1$ of size $|{\cal U}_1| \geq U$ the trusted party needs to generate $T$ random symbols in $\mathbb{F}_q^{\frac{d}{U-T}}$, which account to a total amount of randomness that increases exponentially with $N$. In sharp contrast, \scheme does not require a trusted third party, and each user generates \emph{locally} a set of $T$ random symbols. It significantly improves the practicality of \scheme to maintain model security, and further reduces the total amount of needed randomness to scale linearly with $N$. Consequently, the local offline storage of each user in \scheme scales linearly with $N$, as opposed to scaling exponentially in~\cite{zhao2021information}. We compare the amount of generated randomness and the offline storage cost between the scheme in~\cite{zhao2021information} and \scheme in Table~\ref{table:compare_hua}.
\section{Experimental Details}\label{appendix:exp}
\begin{figure}[t!]
    \centering
    \subfigure[Non-overlapped]{\label{fig:runtime_Logistic_NonOverlap_varDropout}
    \includegraphics[width=.45\textwidth]{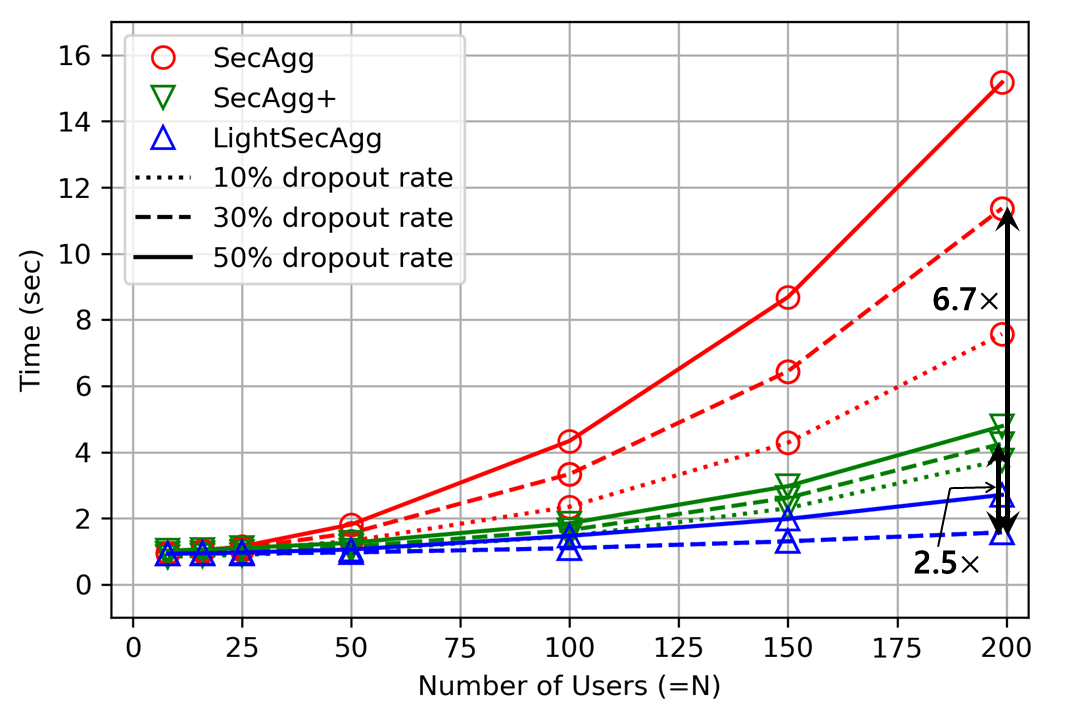}
    }
    \subfigure[Overlapped]{\label{fig:runtime_Logistic_Overlap_varDropout}
    \includegraphics[width=.45\textwidth]{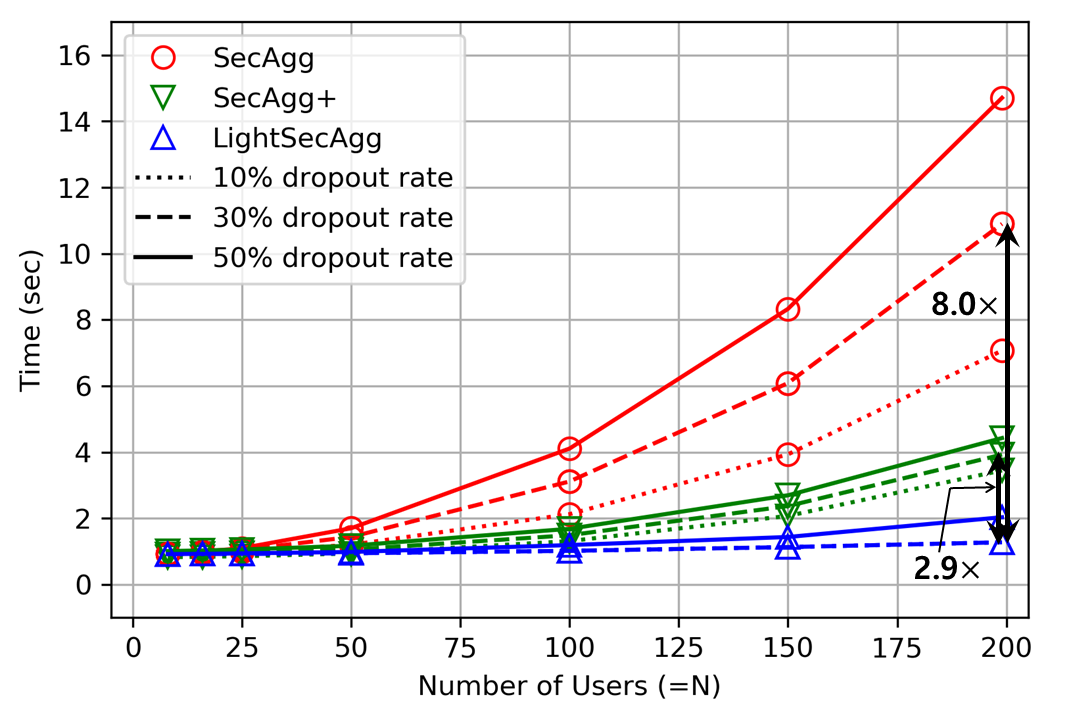}
    }
    \vspace{-6pt}
\caption{Total running time of \scheme versus the state-of-the-art protocols (\google \cite{bonawitz2017practical} and \googlep \cite{bell2020secure}) to train logistic regression on the MNIST~\cite{lecun1998gradient} with an increasing number of users, for various dropout rate.}
\label{fig:runtime_Logistic_varDropout}
\end{figure}

\begin{figure}[t!]
    \centering
    \subfigure[Non-overlapped]{\label{fig:runtime_MobileNet_NonOverlap_varDropout}
    \includegraphics[width=.45\textwidth]{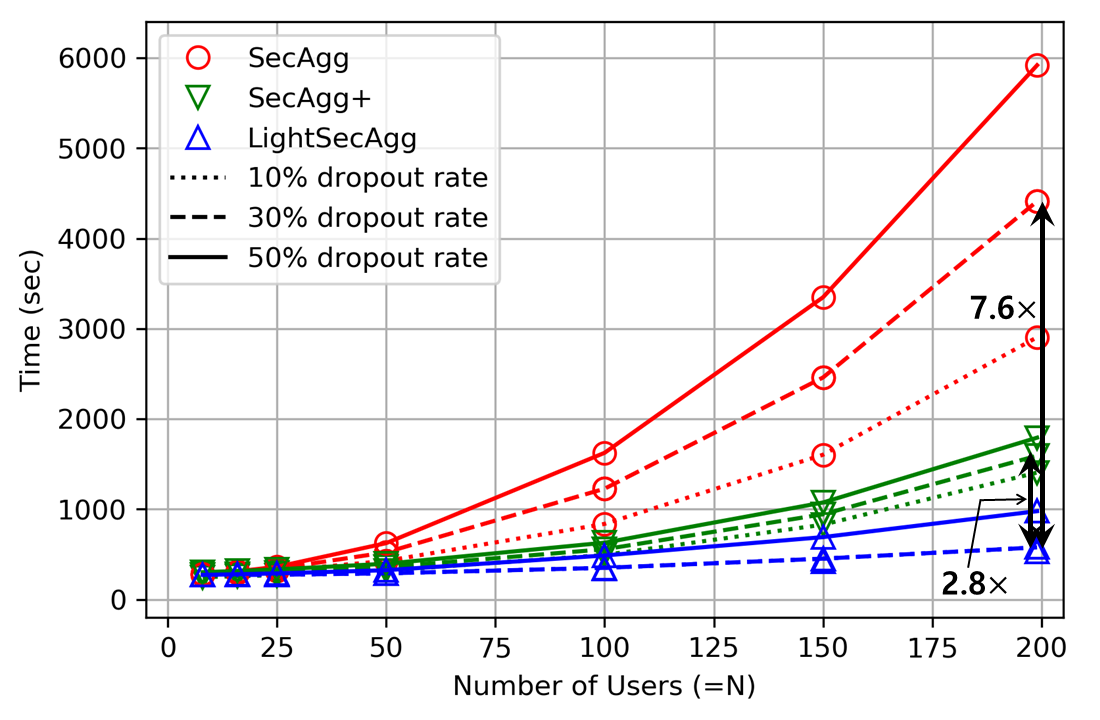}
    }
    \subfigure[Overlapped]{\label{fig:runtime_MobileNet_Overlap_varDropout}
    \includegraphics[width=.45\textwidth]{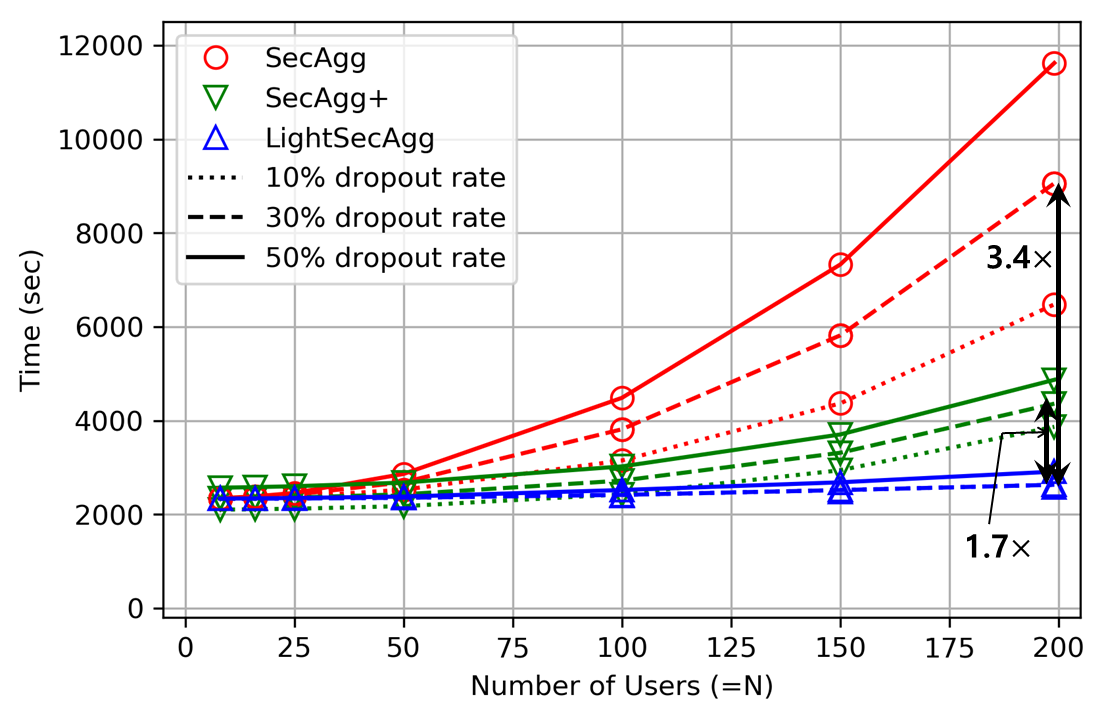}
    }
    \vspace{-6pt}
\caption{Total running time of \scheme versus the state-of-the-art protocols (\google \cite{bonawitz2017practical} and \googlep \cite{bell2020secure}) to train MobileNetV3~\cite{howard2019searching} on the CIFAR-10~\cite{krizhevsky2009learning} with an increasing number of users, for various dropout rate.}
\label{fig:runtime_MobileNet_varDropout}
\end{figure}

\begin{figure}[t!]
    \centering
    \subfigure[Non-overlapped]{\label{fig:runtime_EfficientNet_NonOverlap_varDropout}
    \includegraphics[width=.45\textwidth]{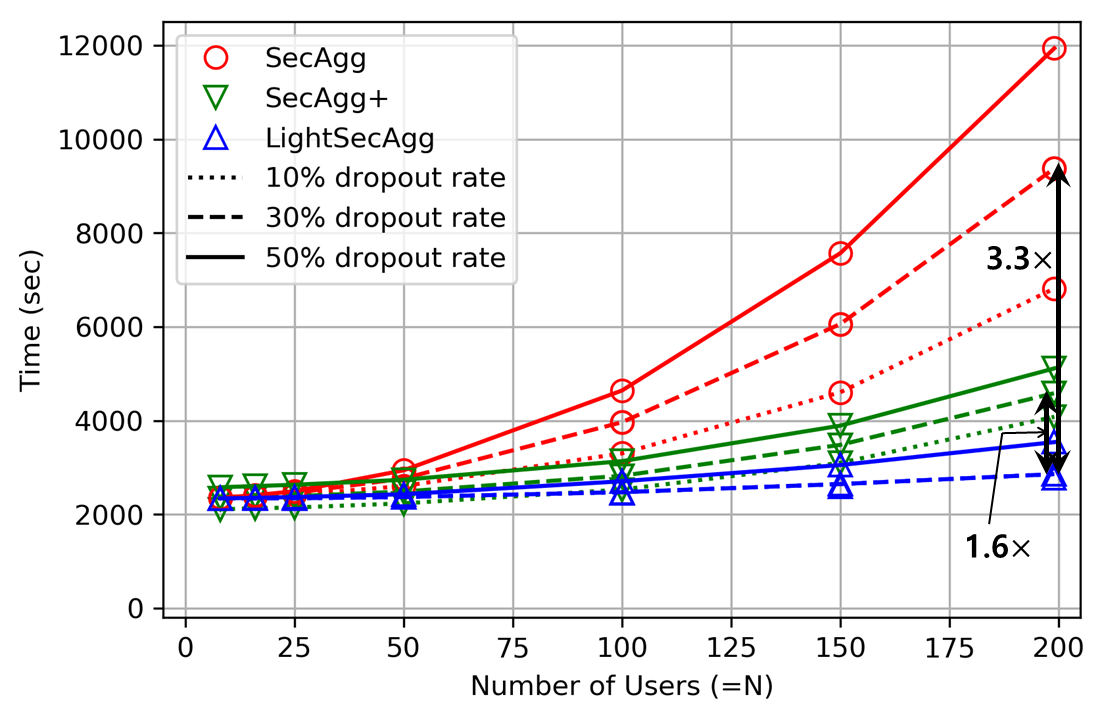}
    }
    \subfigure[Overlapped]{\label{fig:runtime_EfficientNet_Overlap_varDropout}
    \includegraphics[width=.45\textwidth]{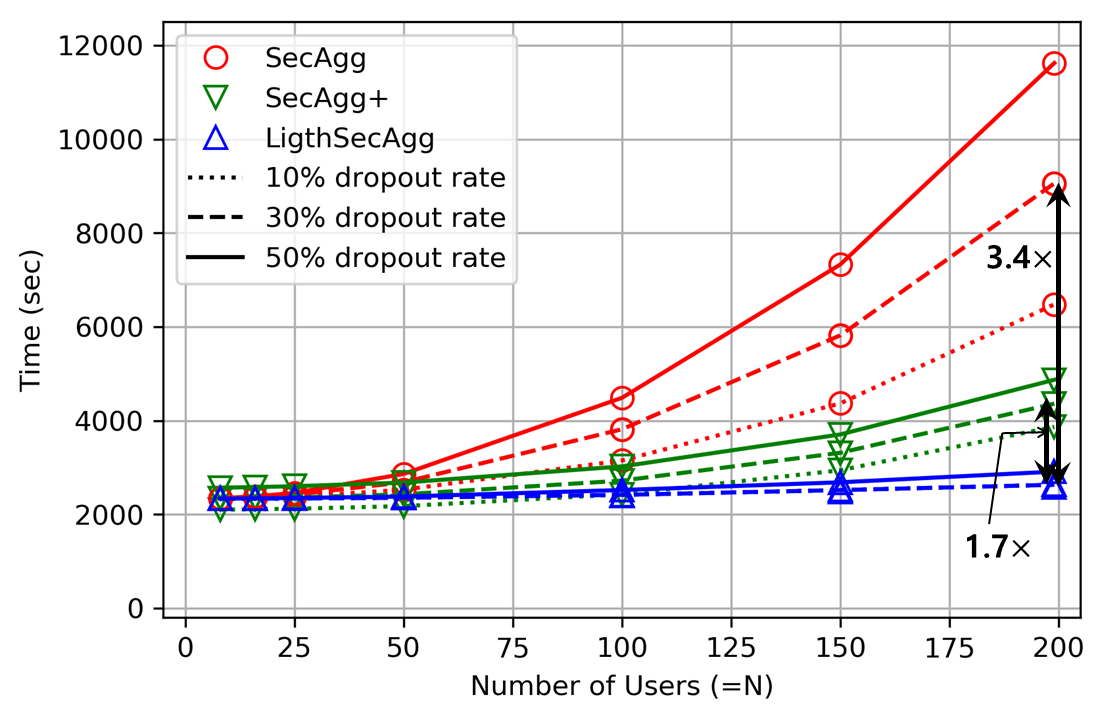}
    }
    \vspace{-6pt}
\caption{Total running time of \scheme versus the state-of-the-art protocols (\google \cite{bonawitz2017practical} and \googlep~\cite{bell2020secure}) to train EfficientNet-B0~\cite{tan2019efficientnet} on the GLD23k~\cite{weyand2020google} with an increasing number of users, for various dropout rate.}
\label{fig:runtime_EfficientNet_varDropout}
\end{figure}

In this section, we provide experimental details of Section \ref{sec:experiment}.
Aside from the results of training CNN~\cite{mcmahan2017communication} on the FEMNIST dataset~\cite{caldas2018leaf} as shown in Figure~\ref{fig:runtime_CNN_varDropout}, we also demonstrate the total running time of \scheme versus two baseline protocols \google \cite{bonawitz2017practical} and \googlep \cite{bell2020secure} to train logistic regression on the MNIST dataset~\cite{lecun1998gradient}, MobileNetV3~\cite{howard2019searching} on the CIFAR-10 dataset~\cite{krizhevsky2009learning}, and EfficientNet-B0~\cite{tan2019efficientnet} on the GLD23k dataset~\cite{weyand2020google} in Figure \ref{fig:runtime_Logistic_varDropout}, Figure \ref{fig:runtime_MobileNet_varDropout}, and Figure \ref{fig:runtime_EfficientNet_varDropout}, respectively. For all considered FL training tasks, each user locally trains its model with $E=5$ local epochs, before masking and uploading its model. We can observe that \scheme provides significant speedup for all considered FL training tasks in the running time over \google and \googlep. 

\clearpage
\section{Proof of Lemma~\ref{lemma}}\label{appendix:lemma}
We show that for an arbitrary set of colluding users $\mathcal{T}$ of size $T$, we have
\begin{align}
    I(\{\mathbf{z}_i\}_{i\in[N]\backslash \mathcal{T}};\{\mathbf{z}_i\}_{i\in \mathcal{T}},\{[\mathbf{\tilde{z}}_j]_i\}_{j\in [N], i \in \mathcal{T}}) = 0. \label{eq:privacy_lemma}
\end{align}
The $T$-private MDS matrix used in \scheme guarantees $I(\mathbf{z}_i;\{[\mathbf{\tilde{z}}_i]_j\}_{j \in \mathcal{T}}) = 0$. Thus, 
\begin{align}
    &  I(\{\mathbf{z}_i\}_{i\in[N]\backslash \mathcal{T}};\{\mathbf{z}_i\}_{i\in \mathcal{T}},\{[\mathbf{\tilde{z}}_j]_i\}_{j\in [N], i \in \mathcal{T}})\\
    = & H(\{\mathbf{z}_i\}_{i\in\mathcal{T}},\{[\mathbf{\tilde{z}}_j]_i\}_{j\in [N], i \in \mathcal{T}}) -H(\{\mathbf{z}_i\}_{i\in \mathcal{T}},\{[\mathbf{\tilde{z}}_j]_i\}_{j\in [N], i \in \mathcal{T}}|\{\mathbf{z}_i\}_{i \in [N]\backslash \mathcal{T}}) \\
    = &  H(\{\mathbf{z}_i\}_{i\in\mathcal{T}},\{[\mathbf{\tilde{z}}_j]_i\}_{j\in [N], i \in \mathcal{T}}) - H(\{\mathbf{z}_i\}_{i\in \mathcal{T}}|\{\mathbf{z}_i\}_{i \in [N]\backslash \mathcal{T}}) - H(\{[\mathbf{\tilde{z}}_j]_i\}_{j\in [N], i \in \mathcal{T}}|\{\mathbf{z}_i\}_{i \in [N]}) \label{eq:lemma1}\\
    = & H(\{\mathbf{z}_i\}_{i\in\mathcal{T}},\{[\mathbf{\tilde{z}}_j]_i\}_{j\in [N], i \in \mathcal{T}}) - H(\{\mathbf{z}_i\}_{i\in \mathcal{T}}) - H(\{[\mathbf{\tilde{z}}_j]_i\}_{j\in [N], i \in \mathcal{T}}) \label{eq:lemma2}\\
    = & 0 \label{eq:lemma3},
\end{align}
where equation \eqref{eq:lemma1} follows from the chain rule. Equation \eqref{eq:lemma2} follows from the independence of $\mathbf{z}_i$'s and $I(\mathbf{z}_i;\{[\mathbf{\tilde{z}}_i]_j\}_{j \in \mathcal{T}}) = 0$. Equation~\eqref{eq:lemma3} follows from the fact that joint entropy is less than or equal to the sum of the individual entropies, combined with the non-negativity of mutual information.

\section{Application of \scheme to Asynchronous FL}\label{appendix:asyncFL}
In this Appendix, we provide a brief overview of asynchronous FL in Appendix \ref{app:sub:async-overview}.
Then, we illustrate the incompatibility of the conventional secure aggregation protocols, \google and \googlep, with the asynchronous FL in Appendix \ref{app:sub:incompatibility}. 
Later on, in Appendix \ref{app:sub:async-lightsecagg}, we demonstrate how \scheme can be applied to the asynchronous FL setting to protect the privacy of individual updates. 

\subsection{General Description of Asynchronous FL}\label{app:sub:async-overview}
We consider the general asynchronous FL setting where the updates of the users are not synchronized while the goal is the same as synchronous FL, to collaboratively learn a  global model $\mathbf{x} \in \mathbb R^d$, using the local datasets of $N$ users without sharing them.
This problem is formulated as minimizing a global loss function as follows
\begin{equation}\label{eq:objective_fnc} 
    \min_{\mathbf{x} \in \mathbb R^d} 
    F(\mathbf{x}) = \sum_{i=1}^N p_i F_i (\mathbf{x}), 
\end{equation} 
where $F_i$ is the local loss function of user $i \in [N]$ and $p_i \geq 0$ are the weight parameters that indicate the relative impact of the users and are selected such that $\sum_{i=1}^{N} p_i=1$.
This problem is solved iteratively in asynchronous FL.

At round $t$, each user locally trains the model by carrying out $E\geq1$ local SGD steps.
When the local update is done, user $i$ sends the difference between the downloaded global model and updated local model to the server. 
The local update of user $i$ sent to the server at round $t$ is given by
\begin{equation}\label{eq:local_update}
    {\Delta}^{(t;t_i)}_i = \mathbf{x}^{(t_i)} - \mathbf{x}^{(E;t_i)}_i,
\end{equation}
where $t_i$ is the latest round index when the global model is downloaded by user $i$ and $t$ is the round index when the local update is sent to the server, hence the staleness of user $i$ is given by $\tau_i = t- t_i$. $\mathbf{x}^{(E;t_i)}_i$ denotes the local model after $E$ local SGD steps and the local model at user $i$ is updated as  
\begin{equation}\label{eq:onestep_localSGD}
    \mathbf{x}^{(e;t_i)}_i = \mathbf{x}^{(e-1;t_i)}_i - \eta_l g_i(\mathbf{x}^{(e-1;t_i)}_i;\xi_i)
\end{equation}
for $e=1,\ldots,E$, where $\mathbf{x}^{(0;t_i)}_i = \mathbf{x}^{(t_i)}$, $\eta_l$ denotes learning rate of the local updates.
$g_i(\mathbf{x};\xi_i)$ denotes the stochastic gradient with respect to the random sampling $\xi_i$ on user $i$, and we assume $\mathbb{E}_{\xi_i}[g_i(\mathbf{x};\xi_i)] = \nabla F_i(\mathbf{x})$ for all $\mathbf{x}\in\mathbb{R}^d$ where $F_i$ is the local loss function of user $i$ defined in \eqref{eq:objective_fnc}.
When the server receives ${\Delta}^{(t;t_i)}_i$, the global model at the server is updated as  
\begin{equation}\label{eq:global_update}
    \mathbf{x}^{(t+1)} = \mathbf{x}^{(t)} - \frac{\eta_g}{\sum_{i\in\mathcal{S}^{(t)}} s(t-t_i)} \sum_{i\in \mathcal{S}^{(t)}} s(t-t_i){\Delta}^{(t;t_i)}_i,
\end{equation}
where $\mathcal{S}^{(t)}$ is an index set of the users whose local models are sent to the server at round $t$ and $\eta_g$ is the learning rate of the global updates.
$s(\tau)$ is a function that compensates for the staleness satisfying $s(0)=1$ and decreases monotonically as $\tau$ increases. 
There are many functions that satisfy these two properties and we consider a polynomial function $s_{\alpha}(\tau) = (\tau +1)^{-\alpha}$ as it shows similar or better performance than the other functions e.g., Hinge or Constant stale function~\cite{xie2019asynchronous}.

As discussed in Section \ref{subsec:ext_to_async}, we focus  on extending \scheme to the \emph{buffered} asynchronous FL setting of \FedBuff~\cite{nguyen2021federated}, where the server stores the local updates in buffer of size $K$ and updates the global model once the buffer is full.
This is a special case of the general asynchronous FL setting described above, where $|\mathcal{S}^{(t)}|=K$ for all $t$. In principle the same approach for generalizing \scheme can be used in other asynchronous FL settings where $|\mathcal{S}^{(t)}|$ changes over time, however since the convergence of FL in those settings are yet not understood, we do not consider them in the paper.

\subsection{Incompatibility  of \google and \googlep with Asynchronous FL}
\label{app:sub:incompatibility}
As described in Section~\ref{sec:overviews}, \google \cite{bonawitz2017practical} and \googlep~\cite{bell2020secure} are designed for synchronous FL.
At round $t$, each pair of users $i,j\in[N]$ agree on a pairwise random-seed $a_{i,j}^{(t)}$, and generate a random vector by running a PRG based on the random seed of $a_{i,j}^{(t)}$ to mask the local update. 
This additive structure has the unique property that these pairwise random vectors cancel out when the server aggregates the masked models because user $i(<j)$ adds $\mathrm{PRG}(a_{i,j}^{(t)})$ to $\mathbf{x}_i^{(t)}$ and user $j(>i)$ subtracts $\mathrm{PRG}(a_{i,j}^{(t)})$ from $\mathbf{x}_j^{(t)}$.

In asynchronous FL, however, the cancellation of the pairwise random masks based on the key agreement protocol is not guaranteed due to the mismatch in staleness between the users. Specifically, at round $t$, user $i\in\mathcal{S}^{(t)}$ sends the masked model $\mathbf{y}_i^{(t;t_i)}$ to the server that is given by
\begin{equation} \small
    \mathbf{y}_i^{(t;t_i)} = {\Delta}^{(t;t_i)}_i + \mathrm{PRG}\left(b_i^{(t_i)}\right)+\sum_{j: i<j} \mathrm{PRG}\left(a^{(t_i)}_{i, j}\right)-\sum_{j: i>j} \mathrm{PRG}\left(a^{(t_i)}_{j, i}\right),
\end{equation}
where ${\Delta}^{(t;t_i)}_i$ is the local update defined in \eqref{eq:local_update}.
When $t_i \neq t_j$, the pairwise random vectors in $\mathbf{y}_i^{(t;t_i)}$ and $\mathbf{y}_j^{(t;t_j)}$ are not canceled out as $a^{(t_i)}_{i, j} \neq a^{(t_j)}_{i, j}$.
We note that the identity of the staleness of each user is not known a priori, hence each pair of users cannot use the same pairwise random-seed.

\subsection{Asynchronous \scheme}\label{app:sub:async-lightsecagg}

We now demonstrate how \scheme can be applied to the asynchronous FL setting where the server stores each local update in a buffer of size $K$ and updates the global model by aggregating the stored updates when the buffer is full. Our key intuition is to encode the local masks in a way that the server can recover the aggregate of masks from the encoded masks via a one-shot computation even though the masks are generated in different training rounds.
The asynchronous \scheme protocol also consists of three phases with three design parameters $D, T, U$ which are defined in the same way as the synchronous \scheme.

Synchronous and asynchronous \scheme have two key differences: (1) In asynchronous FL, the users share the encoded masks with the time stamp in the first phase to figure out which encoded masks should be aggregated for the reconstruction of aggregate of masks in the third phase. Due to the commutative property of coding and addition, the server can reconstruct the aggregate of masks even though the masks are generated in different training rounds; (2) In asynchronous FL, the server compensates the staleness of the local updates. This is challenging as this compensation should be carried out over the masked model in the finite field to provide the privacy guarantee while the conventional compensation functions have real numbers as outputs \cite{xie2019asynchronous, nguyen2021federated}.

We now describe the three phases in detail. 

\subsubsection{Offline Encoding and Sharing of Local Masks}\label{subsubsec:BASecAgg_firstphase} 
User $i$ generates $\mathbf{z}_i^{(t_i)}$ uniformly at random from the finite field $\mathbb{F}^d_q$, where $t_i$ is the global round index when user $i$ downloads the global model from the server. The mask $\mathbf{z}_i^{(t_i)}$ is partitioned into $U-T$ sub-masks denoted by 
$[\mathbf z^{(t_i)}_i]_1, \cdots, [\mathbf z^{(t_i)}_i]_{U-T}$, where $U$ denotes the targeted number of surviving users and $N-D\geq U \geq T$. User $i$ also selects another $T$ random masks denoted by $[\mathbf n^{(t_i)}_i]_{U-T+1}, \cdots, [\mathbf n^{(t_i)}_i]_{U}$. These $U$ partitions $[\mathbf z^{(t_i)}_i]_1, \cdots, [\mathbf z^{(t_i)}_i]_{U-T}, [\mathbf n^{(t_i)}_i]_{U-T+1}, \cdots, [\mathbf n^{(t_i)}_i]_{U}$ are then encoded through an $(N, U)$ Maximum Distance Separable (MDS) code as follows 
\begin{align}\label{eq:encoding}
   [\widetilde{\mathbf z}^{(t_i)}_i]_j= \left([\mathbf z_i^{(t_i)}]_1, \cdots, [\mathbf z^{(t_i)}_i]_{U-T}, [\mathbf{n}^{(t_i)}_i]_{U-T+1}, \cdots, [\mathbf{n}^{(t_i)}_i]_{U} \right) \mathbf W_j,
\end{align}
where $\mathbf W_j$ is the Vandermonde matrix defined in \eqref{eq:def_z_tilde}.
User $i$ sends $[\widetilde{\mathbf z}^{(t_i)}_i]_j$ to user $j \in [N] \setminus \{i\}$.
At the end of this phase, each user $i\in[N]$ has $[\widetilde{\mathbf z}^{(t_j)}_j]_i$ from $j\in[N]$.


\subsubsection{Training, Quantizing, Masking, and Uploading of Local Updates}\label{subsubsec:BASecAgg_secondphase}
Each user $i$ trains the local model as in \eqref{eq:local_update} and \eqref{eq:onestep_localSGD}. 
User $i$ quantizes its local update ${\Delta}^{(t;t_i)}_i$ from the domain of real numbers to the finite field $\mathbb{F}_q$ as masking and MDS encoding are carried out in the finite field to provide information-theoretic privacy.
The field size $q$ is assumed to be large enough to avoid any wrap-around during secure aggregation.

The quantization is a challenging task as it should be performed in a way to ensure the convergence of the global model. Moreover, the quantization should allow the representation of negative integers in the finite field, and enable computations to be carried out in the quantized domain. Therefore, we cannot utilize well-known gradient quantization techniques such as in \cite{alistarh2017qsgd}, which represents the sign of a negative number separately from its magnitude.  
\scheme addresses this challenge with a simple stochastic quantization strategy combined with the two's complement representation as described subsequently. 
For any positive integer $c\geq1$, we first define a stochastic rounding function as
\begin{equation}\label{eq:sto_round}
    Q_c(x) = 
    \left\{
    \begin{array}{ll}
          \frac{\lfloor cx \rfloor}{c}   & \text{with prob. } 1-(cx-\lfloor cx \rfloor)\\
          \frac{\lfloor cx \rfloor+1}{c} & \text{with prob. } cx-\lfloor cx \rfloor,
    \end{array} 
    \right. 
\end{equation}
where $\floor{x}$ is the largest integer less than or equal to $x$, and this rounding function is unbiased, i.e., $\mathbb{E}_Q[Q_c(x)]=x$. The parameter $c$ is a design parameter to determine the number of quantization levels. 
The variance of $Q_c(x)$ decreases as the value of $c$ increases.
We then define the quantized update
\begin{equation}\label{eq:def_w_bar}
    \overline{\Delta}^{(t;t_i)}_i := \phi\left({c_l}\cdot Q_{c_l}\left({\Delta}^{(t;t_i)}_i\right)\right),
\end{equation}
where the function $Q_c$ from~\eqref{eq:sto_round} is carried out element-wise, and $c_l$ is a positive integer parameter to determine the quantization level of the local updates.   
The mapping function $\phi:\mathbb{R}\rightarrow\mathbb{F}_q$ is defined to represent a negative integer in the finite field by using the two's complement representation, 
\begin{equation}\label{eq:phi} 
    \phi(x) =
    \left\{
    \begin{array}{ll}
          x & \text{if } x \geq 0\\
          q+x & \text{if } x<0.
    \end{array} 
    \right. 
\end{equation}
To protect the privacy of the local updates, user $i$ masks the quantized update $\overline{\Delta}^{(t;t_i)}_i$ in \eqref{eq:def_w_bar} as
\begin{equation}\label{eq:masked_update}
    \widetilde{\Delta}^{(t;t_i)}_i = \overline{\Delta}^{(t;t_i)}_i + \mathbf{z}_i^{(t_i)},
\end{equation}
and sends the pair of $\left\{\widetilde{\Delta}^{(t;t_i)}_i, t_i \right\}$ to the server.
The local round index $t_i$ is used in two cases: (1) when the server identifies the staleness of each local update and compensates it, and (2) when the users aggregate the encoded masks for one-shot recovery, which will be explained in Section \ref{subsubsec:BASecAgg_thirdphase}.

\subsubsection{One-shot Aggregate-update Recovery and Global Model Update}\label{subsubsec:BASecAgg_thirdphase} 
The server stores $\widetilde{\Delta}^{(t;t_i)}_i$ in the buffer, and when the buffer of size $K$ is full, the server aggregates the $K$ masked local updates. In this phase, the server intends to recover \begin{equation}
    \sum_{i\in\mathcal{S}^{(t)}} s(t-t_i) {\Delta}^{(t;t_i)}_i,    
\end{equation}
where ${\Delta}^{(t;t_i)}_i$ is the local update in the real domain defined in \eqref{eq:local_update}, $\mathcal{S}^{(t)}$ ($\left| \mathcal{S}^{(t)} \right|=K$) is the index set of users whose local updates are stored in the buffer and aggregated by the server at round $t$, 
and $s(\tau)$ is the staleness function defined in \eqref{eq:global_update}.
To do so, the first step is to reconstruct $\sum_{i\in\mathcal{S}^{(t)}} s(t-t_i)\mathbf{z}_i^{(t_i)}$. This is challenging as the decoding should be performed in the finite field, but the value of $s(\tau)$ is a real number.
To address this problem, we introduce a quantized staleness function $\overline{s}:\{0,1,\ldots,\}\rightarrow \mathbb{F}_q$,
\begin{equation}\label{eq:quantized_stale_function}
    \overline{s}_{c_g}(\tau) = c_g Q_{c_g}\left( s(\tau) \right),
\end{equation}
where $Q_c(\cdot)$ is a stochastic rounding function defined in \eqref{eq:sto_round}, and $c_g$ is a positive integer to determine the quantization level of staleness function. 
Then, the server broadcasts information of $\left\{ \mathcal{S}^{(t)}, \left\{ t_i\right\}_{i\in\mathcal{S}^{(t)}}, c_g \right\}$ to all surviving users. 
After identifying the selected users in $\mathcal{S}^{(t)}$, the local round indices $\{t_i\}_{i\in\mathcal{S}^{(t)}}$ and the corresponding staleness, user $j\in[N]$ aggregates its encoded sub-masks $\sum_{i\in\mathcal{S}^{(t)}} \overline{s}_{c_g}(t-t_i) \left[\widetilde{\mathbf z}^{(t_i)}_i\right]_j$
and sends it to the server for the purpose of one-shot recovery. The key difference between the asynchronous \scheme and the synchronous \scheme is that in the asynchronous \scheme, the time stamp $t_i$ for encoded masks $\left[\widetilde{\mathbf z}^{(t_i)}_i\right]_j$ for each $i\in\mathcal{S}^{(t)}$ can be different, hence user $j\in[N]$ must aggregate the encoded mask with the proper round index.
Due to the commutative property of coding and linear operations, each $\sum_{i\in\mathcal{S}^{(t)}} \overline{s}_{c_g}(t-t_i) \left[\widetilde{\mathbf z}^{(t_i)}_i\right]_j$ is an encoded version of $\sum_{i\in\mathcal{S}^{(t)}} \overline{s}_{c_g}(t-t_i) \left[{\mathbf z}^{(t_i)}_i\right]_k$ for $k\in[U-T]$ using the MDS matrix (or Vandermonde matrix) $\mathbf V$ defined in \eqref{eq:encoding}.
Thus, after receiving a set of any $U$ results from surviving users in $\mathcal{U}_2$, where $|\mathcal{U}_2|=U$, the server reconstructs $\sum_{i\in\mathcal{S}^{(t)}} \overline{s}_{c_g}(t-t_i) \left[{\mathbf z}^{(t_i)}_i\right]_k$ for $k\in[U-T]$ via MDS decoding.
By concatenating the $U-T$ aggregated sub-masks $\sum_{i\in\mathcal{S}^{(t)}} \overline{s}_{c_g}(t-t_i) \left[{\mathbf z}^{(t_i)}_i\right]_k$, the server can recover $\sum_{i\in\mathcal{S}^{(t)}} \overline{s}_{c_g}(t-t_i) {\mathbf z}^{(t_i)}_i$.
Finally, the server obtains the desired global update as follows
\begin{equation}
    \mathbf{g}^{(t)} =
    \frac{1}{c_g c_l \sum_{i\in\mathcal{S}^{(t)}} {s}_{c_g}(t-t_i)} \phi^{-1} \left( \sum_{i\in\mathcal{S}^{(t)}} \overline{s}_{c_g}(t-t_i)\widetilde{{\Delta}}^{(t;t_i)}_i - \sum_{i\in\mathcal{S}^{(t)}} \overline{s}_{c_g}(t-t_i) {\mathbf z}^{(t_i)}_i\right),
\end{equation}
where $c_l$ is defined in \eqref{eq:def_w_bar} and $\phi^{-1}:\mathbb{F}_q\rightarrow\mathbb{R}$ is the demapping function defined as follows
\begin{equation}\label{eq:inv_phi}
    {\phi}^{-1}(\overline{x})=
    \left\{
    \begin{array}{ll}
          \overline{x} & \text{if \quad } 0 \leq \overline{x} < \frac{q-1}{2}\\
          \overline{x}-q & \text{if \quad } \frac{q-1}{2} \leq \overline{x} < q.
    \end{array}
    \right.
\end{equation}
Finally, the server updates the global model as $\mathbf{x}^{(t+1)} = \mathbf{x}^{(t)} - \eta_g \mathbf{g}^{(t)}$, which is equivalent to
\begin{equation}\label{eq:global_update_equivalent}
    \mathbf{x}^{(t+1)} = \mathbf{x}^{(t)} - \frac{\eta_g}{\sum_{i\in\mathcal{S}^{(t)}} Q_{c_g}\left(s(t-t_i) \right)} \sum_{i\in \mathcal{S}^{(t)}} Q_{c_g}\left(s(t-t_i) \right) Q_{c_l}\left({\Delta}^{(t;t_i)}_i\right),
\end{equation}
where $Q_{c_l}$ and $Q_{c_g}$ are the stochastic rounding function defined in \eqref{eq:sto_round} with respect to quantization parameters $c_l$ and $c_g$, respectively.

\subsection{Convergence Analysis of Asynchronous \scheme}\label{app:sub:async-convergence}
We now provide the convergence guarantee of asynchronous \scheme in the $L$-smooth and non-convex setting. The prior works mostly focus on the synchronous FL setting, but here we focus on the buffered asynchronous setting. While the convergence analysis in the buffered asynchronous setting has been considered recently in \cite{nguyen2021federated} and the effects of the buffer size and the staleness have been analyzed, \scheme requires quantization to enable secure aggregation without TEEs. Hence, we extend this analysis here by taking the quantization's effect into account.

For simplicity, we consider the constant staleness function $s(\tau)=1$ for all $\tau$ in \eqref{eq:global_update_equivalent}. Then, the global update equation of asynchronous \scheme is given by
\begin{equation}\label{eq:global_update_conv}
    \mathbf{x}^{(t+1)} = \mathbf{x}^{(t)} - \frac{\eta_g}{K} \sum_{i\in \mathcal{S}^{(t)}}  Q_{c_l}\left({\Delta}^{(t;t_i)}_i\right),
\end{equation}
where $Q_{c_l}$ is the stochastic round function defined in \eqref{eq:sto_round}, $c_l$ is the positive constant to determine the quantization level, and ${\Delta}^{(t;t_i)}_i$ is the local update of user $i$ defined in \eqref{eq:local_update}.
We first introduce our assumptions, which are commonly made in analyzing FL algorithms \cite{li2019convergence, nguyen2021federated, reddi2020adaptive, so2021securing}.

\begin{assumption} \label{assumpt:1} (Unbiasedness of local SGD).
For all $i\in[N]$ and $\mathbf{x}\in\mathbb{R}^d$, $\mathbb{E}_{\xi_i}[g_i(\mathbf{x};\xi_i)] = \nabla F_i(\mathbf{x})$ where $g_i(\mathbf{x};\xi_i)$ is the stochastic gradient estimator of user $i$ defined in \eqref{eq:onestep_localSGD}.
\end{assumption}

\begin{assumption} \label{assumpt:2} (Lipschitz gradient). 
$F_1,\ldots,F_N$ in \eqref{eq:objective_fnc} are all $L$-smooth: for all $\mathbf{a},\mathbf{b}\in\mathbb{R}^d$ and $i\in[N]$, 
$\lVert \nabla F_i(\mathbf{a}) - \nabla F_i(\mathbf{b}) \rVert^2 \leq L \lVert \mathbf{a}-\mathbf{b} \rVert^2$.
\end{assumption}

\begin{assumption}\label{assumpt:3} (Bounded variance of local and global gradients). 
The variance of the stochastic gradients at each user is bounded, i.e., $\mathbb{E}_{\xi_i} \left\lVert \nabla g_i(\mathbf{x};\xi_i) - \nabla F_i(\mathbf{x}) \right\rVert^2 \leq \sigma^2_l$ for $i\in [N]$ and $\mathbf{x}\in\mathbb{R}^d$.
For the global loss function $F(\mathbf{x})$ defined in \eqref{eq:objective_fnc}, $\frac{1}{N}\sum_{i=1}^{N} \left\lVert \nabla F_i(\mathbf{x}) - \nabla F(\mathbf{x}) \right\rVert^2 \leq \sigma^2_{g}$ holds.
\end{assumption}
\begin{assumption}\label{assumpt:4}
(Bounded gradient). For all $i\in [N]$, $\lVert\nabla F_i(\mathbf{x})\rVert^2 \leq G$.
\end{assumption}
In addition, we make an assumption on the staleness of the local updates under asynchrony~\cite{nguyen2021federated}.
\begin{assumption}\label{assumpt:5}(Bounded staleness).
For each global round index $t$ and all users $i\in[N]$, the delay $\tau_i^{(t)} = t - t_i$ is not larger than a certain threshold $\tau_{\mathrm{max}}$ where $t_i$ is the latest round index when the global model is downloaded to user $i$.
\end{assumption}

Now, we state our main result for the convergence guarantee of asynchronous \scheme. 

\begin{theorem}\label{thm:convergence}
Selecting the constant learning rates $\eta_l$ and $\eta_g$ such that $\eta_l \eta_g K E \leq \frac{1}{L}$, the global model iterations in \eqref{eq:global_update_conv} achieve the following ergodic convergence rate
\begin{equation}\label{eq:conv_rate}
    \frac{1}{J} \sum_{t=0}^{J-1} \mathbb{E} \left[ \lvert \nabla F(\mathbf{x}^{(t)}) \rvert^2 \right]
    \leq \frac{2F^{*}}{\eta_g \eta_l E K T} + \frac{L \eta_g \eta_l \sigma^2_{c_l}}{2}
    + 3L^2 E^2 \eta_l^2 \left( \eta_g^2 K^2 \tau^2_{\mathrm{max}} \right) \sigma^2,
\end{equation}
where $F^{*}=F(\mathbf{x}^{(0)}) - F(\mathbf{x}^{*})$, $\sigma^2 = G + \sigma_g^2 + \sigma_{c_l}^2$, and $\sigma_{c_l}^2 = \frac{d}{4{c_l}^2} + \sigma^2_l$.
\end{theorem}


The proof of Theorem \ref{thm:convergence} directly follows from the following useful lemma that shows the unbiasedness and bounded variance still hold for the quantized gradient estimator $Q_c(g(\mathbf{x},\xi))$ for any  $\mathbf{x}\in\mathbb{R}^d$.
\begin{lemma}\label{lemma:q_prop}
    For the quantized gradient estimator $Q_c(g(\mathbf{x},\xi))$ with a given vector $\mathbf{x}\in\mathbb{R}^d$
    where $\xi$ is a uniform random variable representing the sample drawn, $g$ is a gradient estimator such that $\mathbb{E}_\xi [g(\mathbf{x},\xi)]=\nabla F(\mathbf{x})$ and $\mathbb{E}_\xi \lVert g(\mathbf{x},\xi) - \nabla F(\mathbf{x}) \rVert^2 \leq \sigma_l^2$, and the stochastic rounding function $Q_c$ is given in~\eqref{eq:sto_round}, the following holds,
    \begin{align}
        \mathbb{E}_{Q,\xi} [Q_c(g(\mathbf{x},\xi))] &= \nabla F(\mathbf{x}) \label{unbiased}\\
        \mathbb{E}_{Q,\xi} \lVert Q_c(g(\mathbf{x},\xi)) - \nabla F(\mathbf{x}) \rVert^2
        &\leq \sigma_c^{2}, \label{variance}
    \end{align}
    where $\sigma_c^2 = \frac{d}{4c^2} + \sigma^2_l$.
\end{lemma}

\begin{proof}
(Unbiasedness). 
Given $Q_c$ in~\eqref{eq:sto_round} and any random variable $x$, it follows that,
\begin{align}
    \mathbb{E}_Q \left[Q_c(x) \mid x\right] 
    =&\; \frac{\lfloor cx \rfloor}{c} \left(1-(cx-\lfloor cx \rfloor)\right) + \frac{(\lfloor cx \rfloor+1)}{c} (cx-\lfloor cx \rfloor) \notag \\
    =&\; x 
\end{align}
from which we obtain the unbiasedness condition in \eqref{unbiased},
\begin{align}
    \mathbb{E}_{Q,\xi} [Q_c(g(\mathbf{x},\xi))] 
    &= \mathbb{E}_{\xi}\big[ \mathbb{E}_{Q}[Q_c(g(\mathbf{x},\xi))\mid g(\mathbf{x},\xi)] \big] \notag \\
    &= \mathbb{E}_{\xi}\big[ g(\mathbf{x},\xi) \big] \notag \\
    &= \nabla F(\mathbf{x}). 
\end{align}

\noindent    
(Bounded variance). 
Next, we observe that,
\begin{align}
    &\mathbb{E}_Q \left[ \big(Q_c(x) - \mathbb{E}_Q[Q_c(x)\mid x]\big)^2 \mid x \right] \notag \\
    &\quad = \left(\frac{\lfloor cx \rfloor}{c} - x\right)^2 (1-(cx-\lfloor cx \rfloor)) 
    + \left(\frac{\lfloor cx \rfloor+1}{c}-x \right)^2(cx-\lfloor cx \rfloor) \notag \\
    &\quad = \frac{1}{c^2}\left(\frac{1}{4} - \left(cx-\lfloor cx \rfloor -\frac{1}{2}\right)^2 \right) \notag \\
    &\quad \leq \frac{1}{4c^2} \label{eq:lem1_ineq1}
\end{align}
from which we obtain the bounded variance condition in \eqref{variance} as follows,
\begin{align}
    &\mathbb{E}_{Q,\xi} \lVert Q_c(g(\mathbf{x},\xi)) - \nabla F(\mathbf{x}) \rVert^2 \notag \\
    &\quad = \mathbb{E}_{\xi}\big[ \mathbb{E}_{Q}[ \lVert Q_c(g(\mathbf{x},\xi)) - \nabla F(\mathbf{x}) \rVert^2 \mid g(\mathbf{x},\xi)] \big] \notag \\
    &\quad \leq \mathbb{E}_{\xi}\big[ \mathbb{E}_{Q}[ \lVert Q_c(g(\mathbf{x},\xi)) - g(\mathbf{x},\xi) \rVert^2 \mid g(\mathbf{x},\xi)] \big] 
     + \mathbb{E}_{\xi}\big[ \mathbb{E}_{Q}[ \lVert g(\mathbf{x},\xi) - \nabla F(\mathbf{x}) \rVert^2 \mid g(\mathbf{x},\xi)] \big] \label{eq:triangle_ineq} \\
    &\quad \leq \frac{d}{4c^2} + \sigma^2_l  \label{eq:lem1_ineq2} \\
    &\quad = \sigma^{2}_c, \notag
\end{align}
where \eqref{eq:triangle_ineq} follows from the triangle inequality and \eqref{eq:lem1_ineq2} follows form~\eqref{eq:lem1_ineq1}.
\end{proof}

Now, the update equation of asynchronous \scheme is equivalent to the update equation of \FedBuff except that \scheme has an additional random source, stochastic quantization $Q_{c_l}$, which also satisfies the unbiasedness and bounded variance. 
One can show the convergence rate of asynchronous \scheme presented in Theorem \ref{thm:convergence} by exchanging $\mathbf{E}_\xi$ and variance-bound $\sigma^2_l$ in \cite{nguyen2021federated} with $\mathbf{E}_{Q_{c_l}, \xi}$ and variance-bound $\sigma^2_{c_l}= \frac{d}{4{c_l}^2} + \sigma^2_l$, respectively.

\begin{remark} \normalfont \label{remark:quantization_var}
Theorem \ref{thm:convergence} shows that convergence rates of asynchronous \scheme and \FedBuff (see Corollary 1 in \cite{nguyen2021federated}) are the same except for the increased variance of the local updates due to the quantization noise in \scheme. 
The amount of the increased variance $\frac{d}{4{c_l}^2}$ in $\sigma_{c_l}^2 = \frac{d}{4{c_l}^2} + \sigma^2_l$ is negligible for large ${c_l}$, which will be demonstrated in our experiments in Appendix \ref{app:sub:async-exp}.
\end{remark}

\subsection{Experiments for Asynchronous \scheme}\label{app:sub:async-exp} 
As described in the previous sections, there is no prior secure aggregation protocol  applicable to  asynchronous FL, and hence we cannot compare the the total running time of \scheme with other baseline protocols, such as \google and \googlep that were considered in synchronous FL.
As such, in our experiments here we instead focus on convergence performance of \scheme compared to the buffered asynchronous FL scheme to highlight the impact of asynchrony and quantization in performance.
We measure the performance in terms of the model accuracy evaluated over the test samples with respect to the global round index $t$.

\noindent {\bf Datasets and network architectures.} We consider an image classification task on the MNIST  \cite{lecun1998gradient} and CIFAR-10 datasets  \cite{krizhevsky2009learning}.
For the MNIST dataset, we train LeNet \cite{lecun1998gradient}. For the  CIFAR-10 dataset, we train the convolutional neural network (CNN) used in \cite{xie2019asynchronous}. 
These network architectures are sufficient for our needs as our goal is to evaluate various schemes, and not to achieve the best accuracy.

\noindent {\bf Setup.} We consider a buffered asynchronous FL setting with $N=100$ users and a single server having the buffer of size $K=10$. For the IID data distribution, the training samples are shuffled and partitioned into $N=100$ users. 
For asynchronous training, we assume the staleness of each user is uniformly distributed over $[0,10]$, i.e., $\tau_{\mathrm{max}}=10$, as used in \cite{xie2019asynchronous}.
We set the field size $q=2^{32}-5$, which is the largest prime within $32$ bits. 
 
\noindent {\bf Implementations.} 
We implement two schemes, \FedBuff and \scheme. 
The key difference between the two schemes is that in \scheme, the local updates are quantized and converted into the finite field to provide privacy of the individual local updates while all operations are carried out over the domain of real numbers in \FedBuff.
For both schemes, to compensate the staleness of the local updates, we employ the two strategies for the weighting function: a constant function $s(\tau)=1$ and a polynomial function $s_\alpha(\tau)=(1+\tau)^{-\alpha}$. 

\begin{figure*}[t!]
\centering
    \subfigure[MNIST dataset.]{\label{fig:CNN_MNIST}
    \includegraphics[scale=0.45]{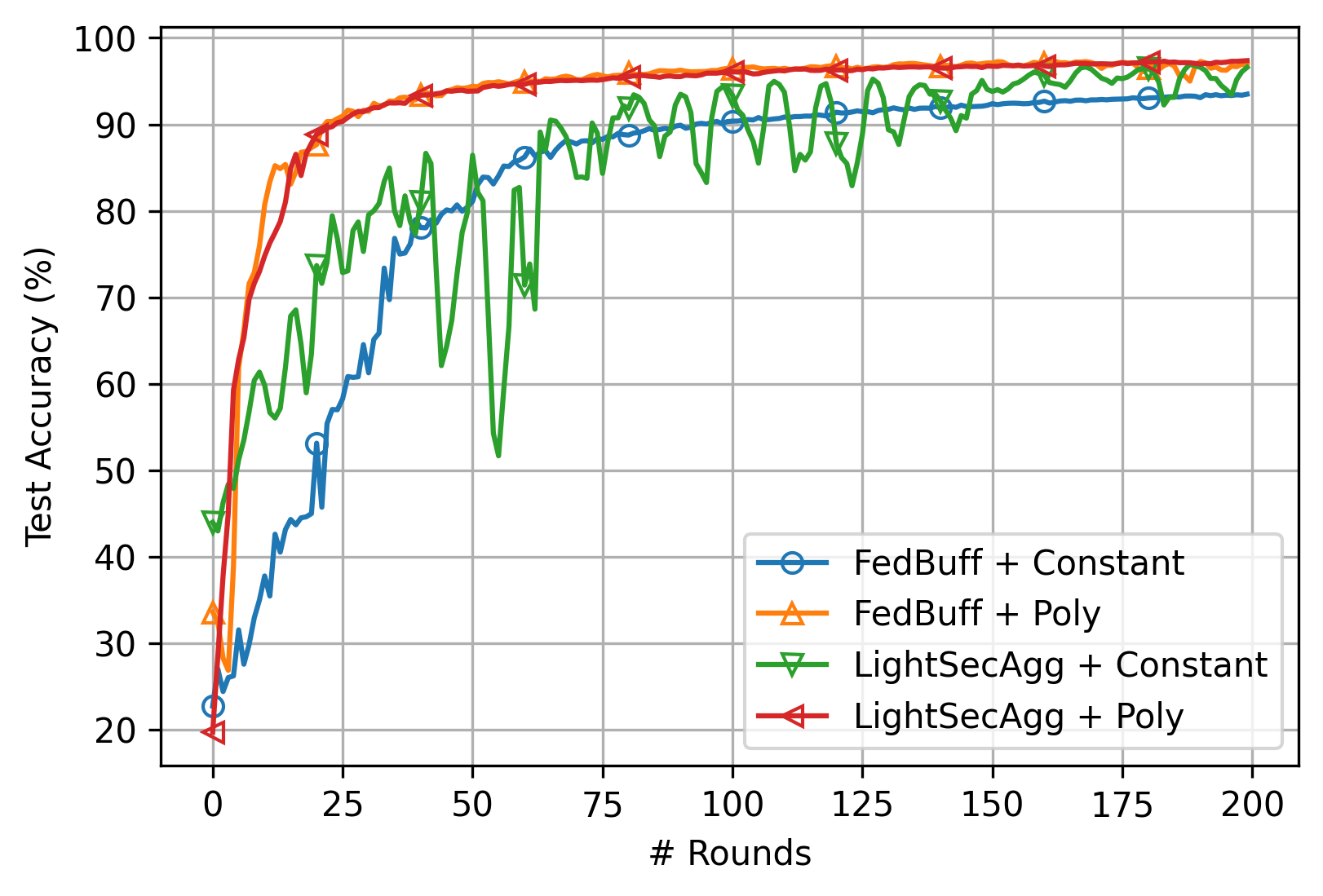}
    }
    \subfigure[CIFAR-$10$ dataset.]{\label{fig:CNN_CIFAR10}
    \includegraphics[scale=0.45]{figures/async_CIFAR10_IID_v1.png}
    }
\caption{\footnotesize Accuracy of asynchronous \scheme and \FedBuff with two strategies for the weighting function to mitigate the staleness: a constant function $s(\tau)=1$ (no compensation) named \emph{Constant}; and a polynomial function $s_\alpha(\tau)=(1+\tau)^{-\alpha}$ named \emph{Poly} where $\alpha=1$.}
\label{fig:accuracy}
\vspace{-0.15cm}
\end{figure*}

\begin{figure*}[t!]
\centering
    \subfigure[MNIST dataset.]{\label{fig:MNIST_diff_q_level}
    \includegraphics[scale=0.45]{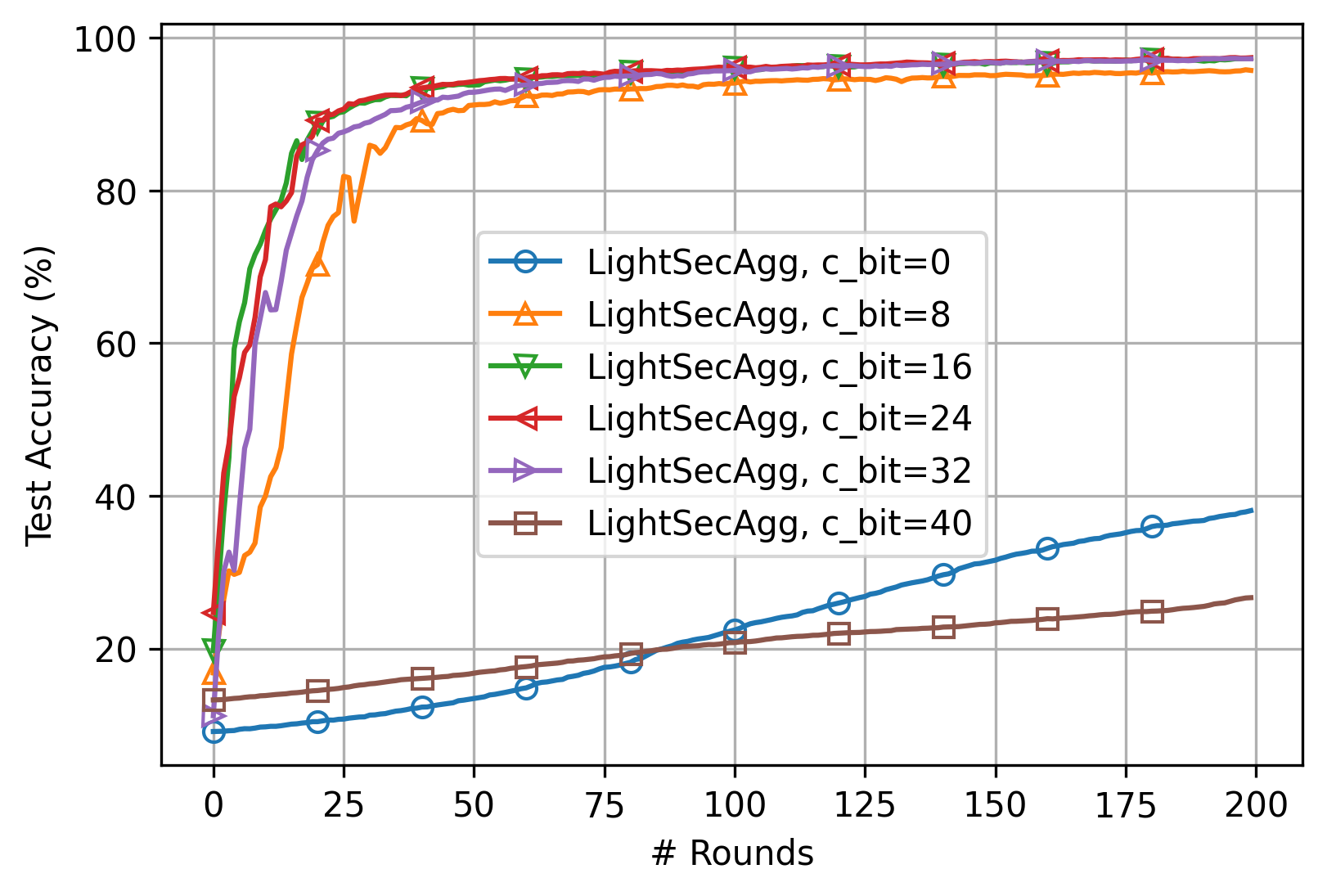}
    }
    \subfigure[CIFAR-$10$ dataset.]{\label{fig:CIFAR10_diff_q_level}
    \includegraphics[scale=0.45]{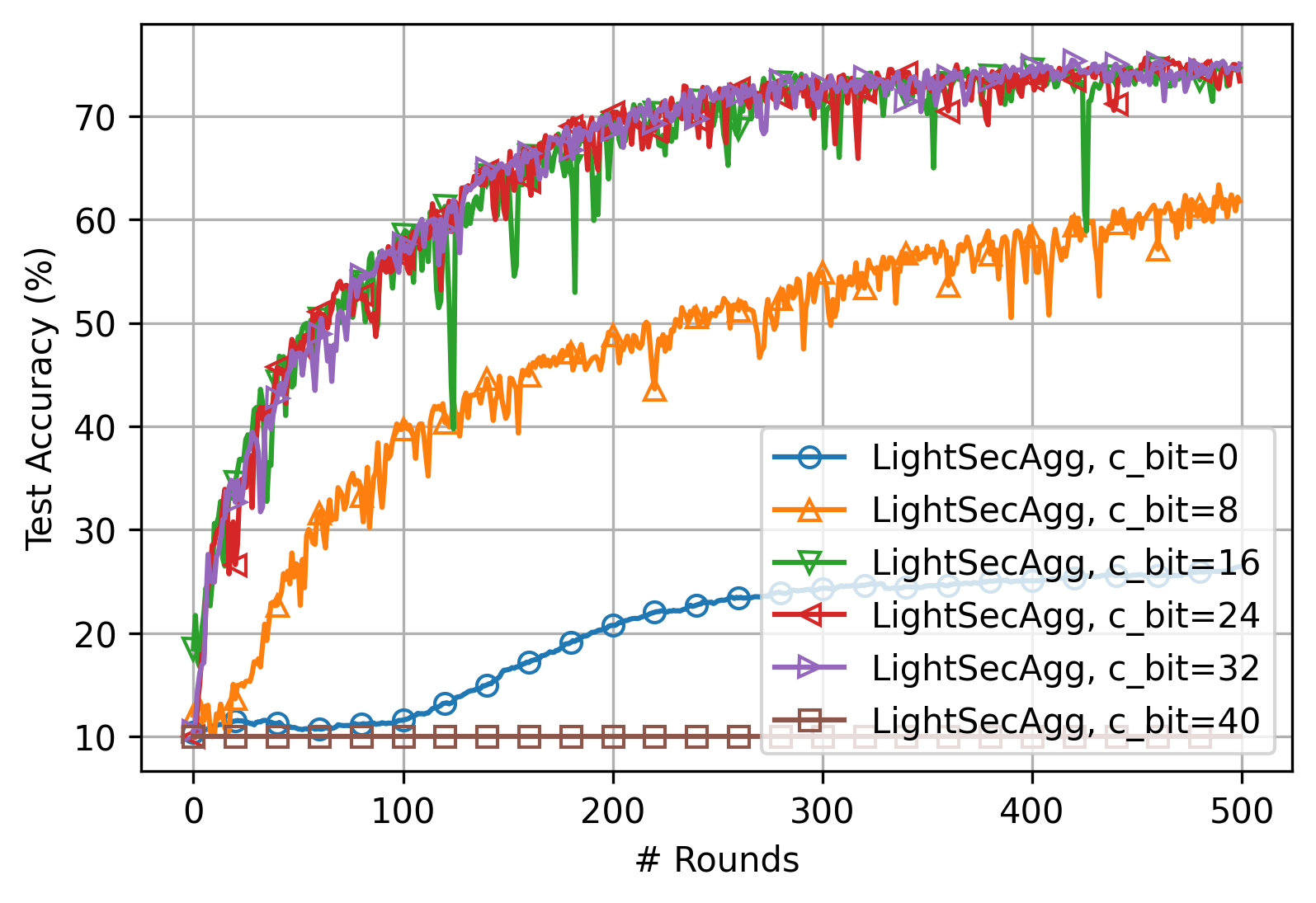}
    }
\vspace{-5 pt}
\caption{\footnotesize Accuracy of asynchronous \scheme and \FedBuff with various values of the quantization parameter $c_l = 2^{c_{bit}}$.}
\label{fig:diff_q_level}
\end{figure*}

\noindent {\bf Empirical results.}
In Figure \ref{fig:CNN_MNIST} and \ref{fig:CNN_CIFAR10}, we demonstrate that \scheme has almost the same performance as \FedBuff on both MNIST and CIFAR-10 datasets, while \scheme includes quantization noise to protect the privacy of individual local updates of users. 
This is because the quantization noise in \scheme is negligible.
To compensate the staleness of the local updates over the finite field in \scheme, we implement the quantized staleness function defined in \eqref{eq:quantized_stale_function} with $c_g=2^6$, which has the same performance in mitigating the staleness as the original staleness function carried out over the domain of real numbers.

\noindent {\bf Performance with various quantization levels.}
To investigate the impact of the quantization, we measure the performance with various values of the quantization parameter $c_l$ on MNIST and CIFAR-10 datasets in Fig. \ref{fig:diff_q_level}. 
We observe that $c_l = 2^{16}$ has the best performance, while a small or a large value of $c_l$ has poor performance. 
This is because the value of $c_l$ provides a trade-off between two sources of quantization noise: 1) the rounding error from the stochastic rounding function defined in \eqref{eq:sto_round} and 2) the wrap-around error when modulo operations are carried out in the finite field. 
When $c_l$ has small value the rounding error is dominant, while the wrap-around error is dominant when $c_l$ has large value. 
To find a proper value of $c_l$, we can utilize the auto-tuning algorithm proposed in \cite{bonawitz2019federated}. 


\end{document}